\documentclass[11pt]{article}
\usepackage[round]{natbib}
\usepackage{graphicx}
\usepackage[margin=0.75in]{geometry}
\usepackage{times}      
\usepackage{amsmath}
\usepackage{amsthm}
\usepackage{amssymb}
\usepackage{subfigure}
\usepackage{authblk}
\usepackage[pdfborder={0 0 0}]{hyperref}
\usepackage{pgf,tikz,pgfplots}
\usetikzlibrary{arrows,automata,fit,positioning}
\newtheorem{theorem}{Theorem}
\newtheorem{proposition}[theorem]{Proposition}
\newtheorem{corollary}[theorem]{Corollary}

\graphicspath{{./icdm_figure/}}
\newcommand{\figref}[1]{{Fig.}~\ref{#1}}
\newcommand{\secref}[1]{{Section}~\ref{#1}}

\newcommand{\tabref}[1]{{Table}~\ref{#1}}
\newcommand{\thref}[1]{{Theorem}~\ref{#1}}

\newcommand{\propref}[1]{{Proposition}~\ref{#1}}

\newcommand{\corref}[1]{{Corollary}~\ref{#1}}

\def\th{{^{th}}}

\def\to{{\,\rightarrow\,}}
\def\kron{\otimes}

\mathchardef\mhyphen="2D

\newcommand{\tvec}[1]{{\text{vec}(#1)}}
\newcommand{\indicator}[1]{ {\mathsf{1}}_{\left[ {#1} \right] }}

\def\half{\frac{1}{2}}
\newcommand{\fracl}[1]{\frac{1}{#1}}

\providecommand{\trans}[1]{{#1^\top}}
\providecommand{\inv}[1]{{#1^{-1}}}
\providecommand{\inner}[2]{{\left\langle#1, #2\right\rangle}}
\providecommand{\innere}[2]{{\trans{#1}#2}}

\newcommand{\tr}[1]{{\mathrm{tr}}\!\left( #1 \right)}
\newcommand{\kl}[2]{{\mathrm{KL}\!\left({#1}\Vert{#2}\right)}}

\newcommand{\vertiii}[1]{{\left\vert\kern-0.25ex\left\vert\kern-0.25ex\left\vert #1
    \right\vert\kern-0.25ex\right\vert\kern-0.25ex\right\vert}}

\newcommand{\normf}[1]{\vertiii{#1}_2}
\newcommand{\normfs}[1]{\normf{#1}^2}
\newcommand{\normt}[1]{\vertiii{#1}_1}

\newcommand{\normh}[2]{\vertiii{#1}_{2 \mhyphen #2}}
\newcommand{\normhs}[2]{\normh{#1}{#2}^2}
\newcommand{\normht}[2]{\vertiii{#1}_{1 \mhyphen #2}}

\newcommand{\gp}[2]{\cG\cP \left(#1, #2\right)}
\newcommand{\gpv}[3]{\cM\cG\cP \left(#1, #2, #3\right)}
\newcommand{\g}[2]{\cN\left(#1, #2\right)}

\newcommand{\ev}[2]{\mathrm{E}_{ #1}\left[ \, #2 \, \right]}

\def\sut{\text{s.t.}}

\newcommand{\D}[2]{\frac{d #1}{d #2}} 
\newcommand{\pd}[2]{\frac{\partial #1}{\partial #2}}

\newcommand{\vect}[1]{{\boldsymbol{#1}}}
\def\eps{\epsilon}

\def\bphi{\vect{\phi}}
\def\bpsi{\vect{\psi}}
\def\btheta{\vect{\theta}}

\def\bgamma{\vect{\gamma}}

\def\bkappa{\vect{\kappa}}

\def\bPsi{\vect{\Psi}}
\def\bPhi{\vect{\Phi}}
\def\bTheta{\vect{\Theta}}
\def\bSigma{\vect{\Sigma}}

\def\bAlpha{\vect{\vA}}

\def\bAlpha{\vect{A}}

\def\sm{{\avec{\vm}}}
\def\sn{{\avec{\vn}}}

\def\ba{{\mathbf{a}}}
\def\bb{{\mathbf{b}}}
\def\bc{{\mathbf{c}}}

\def\bg{{\mathbf{g}}}

\def\bm{{\mathbf{m}}}

\def\bu{{\mathbf{u}}}
\def\bv{{\mathbf{v}}}

\def\bx{{\mathbf{x}}}
\def\by{{\mathbf{y}}}
\def\bz{{\mathbf{z}}}
\def\bA{{\mathbf{A}}}
\def\bB{{\mathbf{B}}}
\def\bC{{\mathbf{C}}}
\def\bD{{\mathbf{D}}}

\def\bG{{\mathbf{G}}}

\def\bI{{\mathbf{I}}}

\def\bL{{\mathbf{L}}}

\def\bP{{\mathbf{P}}}

\def\bS{{\mathbf{S}}}

\def\bU{{\mathbf{U}}}
\def\bV{{\mathbf{V}}}

\def\bX{{\mathbf{X}}}

\def\bZ{{\mathbf{Z}}}

\def\bbE{{\mathbb{E}}}

\def\bbM{{\mathbb{M}}}
\def\bbN{{\mathbb{N}}}

\def\bbR{{\mathbb{R}}}

\def\cB{\mathcal{B}}
\def\cC{\mathcal{C}}
\def\cD{\mathcal{D}}
\def\cE{\mathcal{E}}

\def\cG{\mathcal{G}}
\def\cH{\mathcal{H}}

\def\cM{\mathcal{M}}
\def\cN{\mathcal{N}}
\def\cO{\mathcal{O}}
\def\cP{\mathcal{P}}

\def\cR{\mathcal{R}}
\def\cS{\mathcal{S}}

\def\cX{\mathcal{X}}

\def\sfA{\mathsf{A}}

\def\sfC{\mathsf{C}}

\def\sfL{\mathsf{L}}
\def\sfM{\mathsf{M}}
\def\sfN{\mathsf{N}}

\def\frL{\mathfrak{L}}

\def\sm{_{\mbox{\tiny M}}}
\def\sn{_{\mbox{\tiny N}}}
\def\km{\bC_{\mbox{\tiny M}}}
\def\kn{\bC_{\mbox{\tiny N}}}

\def\gm{\bG_{\mbox{\tiny M}}}
\def\gn{\bG_{\mbox{\tiny N}}}

\def\ckm{\cC_{\mbox{\tiny M}}}
\def\ckn{\cC_{\mbox{\tiny N}}}
\def\cknm{\ckn\kron\ckm}

\def\kt{\bC_{\sfL}}

\def\mn{{m,n}}
\def\smn{_{m,n}}
\def\hm{\cH_{\ckm}}
\def\hn{\cH_{\ckn}}

\def\hk{\cH_{\cC}}

\def\data{\cD}

\def\sigmad{\sigma^2}

\newcommand{\operatorat}[2]{\text{#1}_{@#2}}

\newcommand{\recall}[1]{\operatorat{R}{#1}}
\newcommand{\pres}[1]{\operatorat{P}{#1}}

\def\rmse{\text{RMSE}}

\begin{document}

\title{A Constrained Matrix-Variate Gaussian Process for Transposable Data}

\author[1]{Oluwasanmi Koyejo}
\author[2]{Cheng Lee}
\author[3]{Joydeep Ghosh}

\affil[1]{Imaging Research Center\\
University of Texas at Austin.
\url{sanmi.k@utexas.edu}}
\affil[2]{Biomedical Eng. Dept.\\
University of Texas at Austin.
\url{chlee@utexas.edu}}
\affil[3]{ECE Dept.\\
University of Texas at Austin.
\url{ghosh@ece.utexas.edu}}

\date{}

\maketitle

\begin{abstract}

Transposable data represents interactions among two sets of entities, and are
typically represented as a matrix containing the known interaction values.
Additional side information may consist of feature vectors specific to entities
corresponding to the rows and/or columns of such a  matrix. Further information
may also be available in the form of interactions or hierarchies among entities
along the same mode (axis).
We propose a novel approach for modeling transposable data with missing
interactions given additional side information.
The interactions are modeled as noisy observations from a latent noise free
matrix generated from a matrix-variate Gaussian process.
The construction of row and column covariances using side information provides a
flexible mechanism for specifying a-priori knowledge of the row and column
correlations in the data. Further, the use of such a prior combined with the
side information enables predictions for new rows and columns not observed in
the training data.
In this work, we combine the matrix-variate Gaussian process model with low rank
constraints. The constrained Gaussian process approach is applied to the
prediction of hidden associations between genes and diseases using a small set
of observed associations as well as prior covariances induced by gene-gene
interaction networks and disease ontologies.
The proposed approach is also applied to recommender systems data which involves
predicting the item ratings of users using known associations as well as prior
covariances induced by social networks. We present experimental results that
highlight the performance of constrained matrix-variate Gaussian process as
compared to state of the art approaches in each domain.
\end{abstract}

\section{Introduction}\label{sec:intro}

Transposable data describes relationships between pairs of entities. Such data
can be organized as a matrix, with one set of entities as the rows, the other
set of entities as the columns. In such datasets, both the rows and column of
the matrix are of interest. Transposable data matrices are often sparse, and of
primary interest is the prediction of unobserved matrix entries representing
unknown interactions. In the machine learning community, the modeling of
transposable data is often encountered as multitask learning \citep{stegle11}.
In addition to the matrix, transposable datasets often include features
describing each row entity and each column entity, or graphs describing
relationships between the rows and the columns. These features and graphs can be
useful for improving in-matrix prediction performance and for extending model
predictions outside of the observed matrix, thus alleviating the
\emph{cold-start} problem. In this work, we combine the matrix-variate Gaussian
process model with low rank constraints for the predictive modeling of
transposable data.

In recent years, the matrix variate Gaussian distribution (MV-G) has emerged as
a popular model for transposable data \citep{allen10, allen12} as it compactly
decomposes correlations between the matrix entries into correlations between the
rows, and correlations between the columns. Although the MV-G has been shown to
be effective for modeling matrix data with missing entries, model predictions do
not extend to rows and columns that are unobserved in the training data. One
approach to remedy this deficiency is to replace the MV-G with the nonparametric
matrix-variate Gaussian process (MV-GP) \citep{stegle11}. This is achieved by
replacing the row and column covariance matrices of the MV-G with parameterized
row and column covariance \emph{functions}. Thus, the resulting model can
provide predictions for new rows and columns given features.
The MV-GP may also be described as an extension of the scalar valued Gaussian
process (GP) \citep{rasmussen05}, a popular model for scalar functions, to
vector valued responses. The MV-GP has been applied to link analysis, transfer
learning, collaborative prediction and other multitask learning problems
\citep{yu08, bonilla2008, yan11}. Despite its wide use for transposable data and
multitask learning, the MV-GP does not capture low rank structure.

Rank constraints have become ubiquitous in matrix prediction tasks \citep{yu07,
zhu09, koyejo11, zhou12, koyejo13}. The low rank assumption implies that
matrix-valued parameters of interest can be decomposed as the inner product of
low dimensional factors.
This reduces the degrees of freedom in the matrix model and can improve the
parsimony of the results. Recent theoretical \citep{candes08} and empirical
\citep{koren09} results have provided additional motivation for the low rank
approach. The low rank assumption is also motivated by computational concerns.
Consider the computational requirements of a full matrix regression model such
as a Gaussian process regression \citep{rasmussen05}. Here, the memory
requirements scale quadratically with data size, and na\"{i}ve inference via
using a matrix inverse scales cubically with data size \citep{alvarez11}. In
contrast, training low rank models can scale linearly with the data size and
quadratically with the underlying matrix rank (using the factor representation).
Further, efficient optimization methods have been proposed \citep{koren09,
dudik12}.

We propose a novel constrained Bayesian inference approach that combines the
flexibility and extensibility of the matrix-variate Gaussian process with the
parsimony and empirical performance of low rank models. Constrained Bayesian
inference \citep{koyejo13} is a principled approach for enforcing expectation
constraints on the Bayesian inference procedure. It is a useful approach for
probabilistic inference when the problem of interest requires constraints that
are difficult to capture using standard prior distributions alone. Examples
include linear inequality constraints \citep{gelfand92} and margin constraints
\citep{zhu12}. To enforce these restrictions, constrained Bayesian inference
represents the Bayesian inference procedure as a constrained relative entropy
minimization problem. The resulting optimization problem can often be reduced to
constrained parameter estimation and solved using standard optimization
theoretic techniques.

The main contributions of this paper are as follows:
\begin{itemize}
  \item We propose a novel approach for capturing the low rank
  characteristics of transposable data by combining the matrix-variate Gaussian
  process prior with constrained Bayesian inference subject to nuclear norm
  constraints.
  \item We show that (i) the distribution that solves the constrained Bayesian
  inference problem is a Gaussian process, (ii) its inference can be reduced to
  the estimation of a finite set of parameters, and (iii) the resulting optimization
  problem is strongly convex.
  \item We evaluate the proposed model empirically and show that it
  performs as well as (or better than) the state of the art domain
  specific models for disease-gene association prediction with gene network and
  disease ontology side information and recommender systems with social network
  side information.
\end{itemize}
We begin by discussing relevant background on the matrix-variate Gaussian
process and nuclear norm constraints for matrix-variate functions in
\secref{sec:background}. We introduce the concept of constrained inference
in \secref{sec:constrain} and apply it to the matrix-variate Gaussian process to
compute a low rank prediction (\secref{sec:nucleargp}). We present the
empirical performance of the proposed model compared to state of the art domain
specific models for transposable data in the disease-gene association domain
(\secref{sec:disease-gene}) and the recommender systems domain
(\secref{sec:recsys}). Finally, we conclude in \secref{sec:conclude}.

\section{Background}\label{sec:background}

This section describes the problem statement (\secref{sec:notation}), and the
main building blocks of our approach - the matrix-variate Gaussian process
(\secref{sec:mvgp}) and constrained Bayesian inference (\secref{sec:constrain}).

\subsection{Preliminaries} \label{sec:prelim}
We denote vectors by bold lower case e.g. $\bx$ and matrices by bold upper case
e.g. $\bX$. Let $\bI_D$ represent the $D \times D$ identity matrix. Given a
matrix $\bA \in \bbR^{P \times Q}$, $\tvec{\bA} \in \bbR^{PQ}$ is the vector
obtained by concatenating columns of $\bA$. Given matrices $\bA \in \bbR^{P
\times Q}$ and $\bB \in \bbR^{P' \times Q'}$, the Kronecker product of $\bA$ and
$\bB$ is denoted as $\bA \kron \bB\in \bbR^{PP'\times QQ'}$. A useful property
is the \emph{Kronecker identity}: $\tvec{\bA \bX \bB} =
(\trans{\bB} \kron \bA) \tvec{\bX}$, where $\bX \in \bbR^{Q \times P'}$ and
$\trans{\bB}$ represents the transpose of $\bB$.
 
Let $\ev{}{\cdot}$ be the \emph{expectation operator} with $\ev{p}{f(z)}
= \int_z p(z)f(z) dz$. The \emph{Kullback-Leibler (KL) divergence} between
densities $q(z)$ and $p(z)$ is given by:
\begin{equation*}
\kl{q(z)}{p(z)} = \ev{q}{\log q(z) - \log p(z)}.
\end{equation*}

Let $\bx \in  \bbR^{P}$ be drawn from a \emph{multivariate Gaussian}
distribution. The density is given as:
\begin{equation*}
\g{\bm}{\bSigma} = \frac{ \exp\left( -\half \text{tr} \left[ \trans{(\bx -
\bm)}\inv{\bSigma}(\bx - \bm) \right]\right) } {(2\pi)^{P/2}|\bSigma|^{P/2}},
\end{equation*} 
where $\bm \in \bbR^{P}$ is the mean vector and $\bSigma \in \bbR^{P \times
P}$ is the covariance matrix. $|\cdot|$ denotes the matrix determinant and
$\tr{\cdot}$ denotes the matrix trace.

\subsection{Transposable Data Notation and Problem Statement}
\label{sec:notation}

Let $\bbM \ni m$ be the index set of rows and $\bbN \ni n$ be the index set of
columns. The index set of observed matrix entries is represented by $\sfL=\{(m,
n)\} \subset \bbM \times \bbN$ with every $l=(m,n)\in \sfL$. We define the
subset of observed rows as the set $\sfM  = \{m\,|\, (m,n) \in \sfL\} \subset
\bbM$ with size $|\sfM|=M$, and the subset of observed columns as the set $\sfN 
= \{n\,|\, (m,n) \in \sfL\} \subset \bbN$ with size $|\sfN|=N$ so $L=|\sfL|\le
M\times N$. Let each entry in the matrix be represented by $y_l$. The observed
subset of the transposable matrix is represented by $\by = \trans{\left[ y_{l_1}
\ldots y_{l_{L}} \right]}$. Our goal is to estimate a predictive model for any
unobserved entries $\{y_{l'}\,|\, l' \notin \sfL\}$ including entries not
observed within the bounds of the training matrix i.e. $\{y_{l'}\,|\, l' \notin
\sfM \times \sfN \}$.

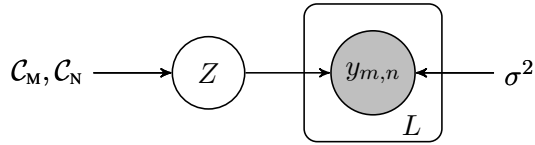
\begin{figure}
\begin{center}
\begin{tikzpicture}[semithick, node distance=4ex, label distance=-0.5ex, ->,
>=stealth', auto] 
\node[state, initial, initial left, initial distance=6ex, initial
text=${\ckm, \ckn}$] [] (z) {$Z$}; 
\node[state, initial, initial right,
initial distance=6ex, initial text=$\sigmad$, fill=black!25,
label=-60:$L$] [right =6.5ex of z] (r) {$y\smn$}; 
\node[draw, inner sep=2ex, rounded corners=1ex, fit=(r)] [] (plate-r)
{};
\path (z) edge (r);n
\end{tikzpicture}
\end{center}
\caption{Plate diagram of the hierarchical matrix-variate Gaussian process
model with i.i.d Gaussian observation noise. $Z(\mn)$ is the hidden noise-free
matrix entry.}\label{fig:mv-gp}
\end{figure}

\subsection{Matrix-variate Gaussian Process for Transposable
Data}\label{sec:mvgp}
The matrix-variate Gaussian process is a doubly indexed stochastic process $\{
Z\smn\}_{m \in \bbM, n \in \bbN}$ where finitely indexed entries are
multivariate Gaussian distributed. As with the scalar Gaussian process
\citep{rasmussen05}, the MV-GP is completely specified by its mean and
covariance functions. We use the notation $\gpv{\phi}{\ckn}{\ckm}$ to denote the
MV-GP with mean function $\phi: \bbM \times \bbN \mapsto \bbR$, row covariance
function $\ckm: \bbM \times \bbM \mapsto \bbR$ and the column covariance
function $\ckn: \bbN \times \bbN \mapsto \bbR$. The covariance function of the
prior MV-GP has a Kronecker product structure \citep{alvarez11}. This form
assumes that the prior covariance between matrix entries can be decomposed as
the product of the row and column covariances. The joint covariance function of
the MV-GP decomposes into product form as $\cC\left((m,n),(m',n')\right) =
\ckm(m,m')\ckn(n,n')$, or equivalently, $\cC = \cknm$. We use the notation
$\gp{\psi}{\cC}$ to denote the scalar valued Gaussian process with mean function
$\psi: \sfL \mapsto \bbR$ and covariance function $\cC: \sfL \times \sfL \mapsto
\bbR$.

Let $Z \sim \gpv{\phi}{\ckm}{\ckn}$, and define the matrix $\bZ \in
\bbR^{M\times N}$ with entries $z\smn = Z(m,n)$ for $\mn \in \sfM \times \sfN$,
$\tvec{\bZ}$ is a distributed as a multivariate Gaussian with mean
$\tvec{\bPhi}$ and covariance matrix $\kn \kron \km$, i.e., $\tvec{\bZ}\sim
\g{\tvec{\bPhi}}{\kn \kron \km}$, where $\phi\smn = \phi(m,n)$, $\bPhi \in
\bbR^{M\times  N}$ is the mean matrix, $\km \in \bbR^{M \times M}$ is the row
covariance matrix and $\kn \in \bbR^{N \times N}$ is the column covariance
matrix. This definition extends to finite subsets $\sfL \subset \bbM \times
\bbN$ that are not complete matrices. For any subset $\sfL$, the vector $\bz =
\left[ z_{l_1} \ldots z_{l_{L}} \right]$ is distributed as $\bz \sim \g{
\bPhi_{\sfL} }{\bC}$ where the vector $\bPhi_{\sfL} =[\phi(1) \ldots \phi(L)]
\in \bbR^L$ are arranged from the entries of the mean matrix corresponding to
the set $l \in \sfL$, and $\bC$ is the covariance matrix evaluated only on pairs
$l,l' \in \sfL \times \sfL$.

The MV-GP is a popular prior distribution for transposable matrix data.
Here we combine it with a Gaussian observation noise model as follows (see
\figref{fig:mv-gp}):
\begin{enumerate}
  \item Draw the function $Z$ from a zero mean MV-GP as $Z \sim
  \gpv{ {0} }{\ckm}{\ckn}$.
  \item Draw observed response independently as $y\smn  \sim
  \g{z\smn}{\sigmad}$ given $z\smn = Z(\mn)$.
\end{enumerate}
The hidden matrix $\bZ \in \bbR^{M \times N}$ with entries $z\smn = Z(\mn)$ may
be interpreted as the latent noise-free matrix. The inference task is to
estimate the posterior distribution $Z|\data$,  where $\data=\{\by, \sfL\}$.
It follows that the posterior distribution is a Gaussian process
\citep{rasmussen05} given by $Z|\data \sim \gp{\phi}{\Sigma}$, with mean and
covariance functions:
\begin{subequations}\label{eq:post_all}
\begin{gather} 
\phi(m, n) = \kt(m,n)\inv{[\bC + \sigmad \bI ]}\by \label{eq:post_mean}
\\
\Sigma\left((m, n), (m',n')\right) = \cC((m,n),(m',n')) - \kt(m,n)\inv{[\bC +
\sigmad \bI ]} \trans{\kt(m,n)}. \label{eq:post_var}
\end{gather}
\end{subequations}
The covariance function $\kt(m,n)$ corresponds to the
sampled covariance matrix between the index $(m,n)$ and all training data
indexes $(m',n')\in \sfL$, $\bC$ is the covariance matrix between all pairs
$(m,n),(m', n') \in \sfL \times \sfL$, and $\bI$ is the $L \times L$ identity
matrix. The closed form follows directly from the definition of a MV-GP as a
scalar GP \citep{rasmussen05} with appropriately vectorized variables. The
computational complexity of applying the GP model scales with the number of
observed samples $L$. Storage of the covariance matrix requires $\cO(L^2)$
memory, and the na\"{i}ve inference requires $\cO(L^3)$ computation.

\subsection{Constrained Bayesian Inference} \label{sec:constrain}

Probabilistic inference involves estimating the distribution of latent variables
given new information such as observed data and constraints. This is often
achieved via Bayes rule. Given the prior distribution of the latent
variables, Bayes rule is a simple formula for computing the latent variable
distribution conditioned on the observed data. However, Bayes rule may be
inadequate when the constraints one seeks to impose on a latent variable
distribution are computationally intractable to enforce by careful selection of
the prior distribution alone. An alternative approach is to enforce these
constraints as part of the inference procedure. While this can be achieved via
rejection sampling and related techniques \citep{gelfand92}, such methods are
computationally intractable for high dimensional variables as a large proportion
of the samples will be discarded. Constrained Bayesian inference via variational
optimization is a useful alternative in such cases. Constrained Bayesian
inference converts the probabilistic inference into an optimization problem,
thus allowing the application of standard optimization techniques.

Let $z$ represent the latent variables and $y$ represent the observations. Bayes
rule can be used to compute the posterior density $p(z|y)$ as:
\begin{equation*}
p(z|y) = \frac{ p(y|z) p(z)}{p(y)}
\end{equation*}
where the conditional density $p(y|z)$ is known as the likelihood, $p(z)$
is the prior density and $p(y)$ is the evidence. An alternative approach
was proposed by \citet{zellner88}, who showed that the Bayesian posterior can
be computed as the solution of the variational optimization problem:
\begin{equation}
p(z|y) =  \underset{q \in \cP}{\arg\min}\;
\kl{q(z)}{p(z)} - \ev{q}{\log p(y|z)}. \label{eq:bayeskl}
\end{equation}
where $\cP = \{q \, | \, \int_z q(z)dz = 1\}$. 

Constrained Bayesian inference \citep{koyejo13} can be used to enforce
additional structure on the posterior density. It involves enforcing
additional constraints on the variational optimization posed in
\eqref{eq:bayeskl}. This paper will focus on expectation constraints applied to
\emph{feature functions} of the latent variables. Given a vector of feature
functions $\bgamma(z)$ and a constraint set $\sfC$, let $\cR_{\sfC} = \{q \in
\cP \, \vert \, \ev{q}{\bgamma(z)} \in \sfC\}$ represent the set of
densities that satisfy the constraint $\ev{q}{\bgamma(z)} \in \sfC$.
Constrained Bayesian inference requires solving one of the following equivalent
variational optimization problems \citep{ganchev10, zhu12, koyejo13}:
\begin{subequations}
\begin{gather}
q_*(z) =  \underset{q \in \cR_{\sfC}}{\arg\min}\; 
  \kl{q(z)}{p(z)} - \ev{q}{\log p(y|z)}.
  \label{eq:conbayes}
\\
q_*(z) = \underset{q \in \cR_{\sfC}}{\arg\min}\; 
  \kl{q(z)}{p(z|y)}.
  \label{eq:conbayes_post}
\end{gather}
\end{subequations}
Thus, the solution is an information projection of the Bayesian posterior
density onto the constraint set $\cC$. Following Zellner, we call $q_*$ the \emph{postdata}
density to distinguish it from the unconstrained Bayesian posterior
density. Further discussion of constrained Bayesian inference is provided
in Appendix \ref{sec:con_proof}.

\section{Related Work} \label{sec:related}

Constrained Bayesian inference is a special case of constrained relative entropy
minimization where some of the constraints are generated from observed data
\citep{koyejo2013a}. Constrained relative entropy minimization and constrained
entropy maximization have been studied in several application domains including
natural language processing \citep{berger96} and ecology \citep{dudik07}.
Applications in the machine learning literature include maximum entropy
discrimination (MED) \citep{jaakkola99}, and other models inspired by MED have
been proposed for combining nonparametric topic models with large margin
constraints for document classification \citep{zhu09} and multitask
classification \citep{zhu11}. Constrained relative entropy models have also been
applied to collaborative filtering \citep{xu2012} and link prediction
\citep{zhu12}. Other work using nonparametric priors \citep{zhu09, zhu11} has
resulted in intractable inference, requiring the application of variational
approximations with tractable assumptions made for the independence structure
and parametric families of the solution. Our work appears to be the first that
uses nonparametric prior distributions without requiring such simplifying
assumptions.
In addition, we consider constraints on the \emph{function space} of the
Gaussian process, which generalize the evaluation based constraints proposed in
prior work i.e. constraints on the entire mean function as opposed to
constraints on the mean of a set of matrix entries.

Factor models such as principal component analysis (PCA) \citep{bishop06} and
its variants are popular methods for extracting information from matrix data.
The standard PCA model can be extended to handle missing data using a Bayesian
approach \citep{bishop06} that marginalizes over the missing data. The Gaussian
process latent variable model (GP-LVM) \citep{lawrence05} was proposed to extend
PCA to model non-linear relationships by replacing the covariance matrix with a
non-linear kernel. This kernel approach has been applied to non-linear matrix
factorization \citep{lawrence09}. The GP-LVM integrates out one of the factors
and estimates the other. The rank of the factor model must be pre-specified in
such models, and is often fixed via expensive cross-validation.
Implementations of Kernel PCA typically capture prior correlations over the rows
\emph{or} the columns, but not both\footnote{The choice to capture either row
or column covariances in PCA and GPLVM is not fundamental to these models i.e.
it is primarily a modeling choice.}. Our proposed model is designed capture
prior correlations simultaneously over the rows and columns via the
matrix-variate Gaussian process prior. Further, the nuclear norm provides an
avenue for automatic (implicit) rank selection.

\begin{figure}
\begin{center}
\begin{tikzpicture}[semithick, node distance=4ex, label distance=-0.5ex, ->,
>=stealth', auto] 
\node[state, initial, initial above, initial distance=6ex, initial
text=$\ckm$, label=-60:$R$] [] (u) {$U^r$}; 
\node[state, initial, initial above,
initial distance=6ex, initial text=$\sigmad$, fill=black!25,
label=-60:$L$] [right=10ex of u] (r) {$y\smn$}; 
\node[state, initial, initial above, initial distance=6ex, initial
text=$\ckn$, label=-60:$R$] [right=10ex of r] (v)
{$V^r$};
\node[draw, inner sep=2ex, rounded corners=1ex, fit=(u)] [] (plate-u)
{};
\node[draw, inner sep=2ex, rounded corners=1ex, fit=(v)] [] (plate-v)
{};
\node[draw, inner sep=2ex, rounded corners=1ex, fit=(r)] [] (plate-r)
{};
\path (u) edge (r);
\path (v) edge (r);
\end{tikzpicture}
\end{center}
\caption{Hierarchical low rank factor Gaussian process.}\label{fig:factor-gp}
\end{figure}
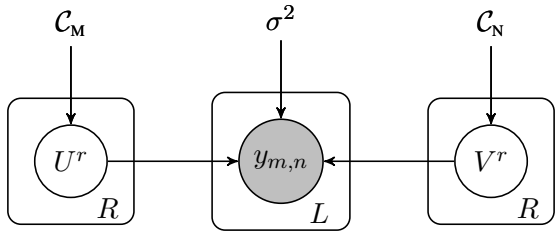

The most common common approach for low rank matrix data modeling in the
Gaussian process literature is the hierarchical low rank factor model. In
particular, the hierarchical low rank factor Gaussian process (factor GP) has
been proposed to capture low rank structure \citep{yu07, zhu09, zhou12}. We
discuss this approach in some detail as it is used as our main baseline. Here,
Gaussian processes are used as the priors for the low dimensional factors. With
a fixed model rank $R$, the generative model for the factor GP is as follows
(see \figref{fig:factor-gp}):
\begin{enumerate}
  \item For each $r\in \{1\ldots R\}$, draw row functions: $U^r \sim \gp{ {0}
  }{\ckm}$. Let $\bu_m \in \bbR^R$ with entries $u^r_m = U^r(m)$.
  \item For each $r\in \{1\ldots R\}$, draw column functions: $V^r \sim \gp{
  {0} }{\ckn}$. Let $\bv_n \in \bbR^R$ with $v^r_n
  = V^r(n)$.
  \item Draw each matrix entry independently: $y\smn  \sim
  \g{\innere{\bu_m}{\bv_n}}{\sigmad} \; \forall \,(m,n)\in \sfL$.
\end{enumerate}
where $\bu_m$ is the $m\th$ row of $\bU=[\bu^1 \ldots \bu^R]\in \bbR^{M\times
R}$, and $\bv_n$ is the $n\th$ row of $\bV=[\bv^1 \ldots \bv^R]\in
\bbR^{N\times R}$.
The maximum-a-posteriori (MAP) estimates of $\bU$ and $\bV$ can be computed as
the solution of the following optimization problem:
\begin{equation}\label{eq:factor_gp}
\underset{\bU, \bV}{\arg\min} \fracl{\sigmad}\sum_{(\mn)\in L}
(y\smn-\innere{\bu_m}{\bv_n})^2 + \tr{\trans{\bU}\inv{\km}\bU} +
\tr{\trans{\bV}\inv{\kn}\bV}
\end{equation}
where $\tr{\bX}$ is the trace of the matrix $\bX$. Statistically, the factor GP may be
interpreted as the sum of rank-one factor matrices. Hence the law of large
numbers can be used to show that the distribution of $Z$ converges to
$\gp{0}{\ckn\kron\ckm}$ as the rank $R \to \infty$ \citep{yu07}.

Despite its success, the factor Gaussian process approach has some deficiencies
when applied for probabilistic inference. First, posterior distributions of
interest are generally intractable. Specifically, neither the joint posterior
distribution of $\{\bU, \bV\}$ nor the distribution of $\bZ=\bU\trans{\bV}$ is
Gaussian, and their posterior distributions are quite challenging to
characterize. As a result, the posterior mean is challenging to compute without
sampling and practitioners often apply the MAP approach. Second order statistics
such as the posterior covariance are also computationally intractable.
Instead, various approximate inference techniques have been applied. A Laplace
approximation was proposed by \citep{yu07} and \citep{zhu09} utilized sampling
techniques. Further, in most cases, the rank must be fixed a-priori. More
recently, Bayesian models for matrix factorization that include a nonparametric
prior for the number of latent factors have been proposed  based on the Indian
buffet process \citep{zhu12b, xu2012} and multiplicative gamma process
\citep{zhang2012}. Inference with these models is generally intractable, and
requires approximations or sampling, which may result in slow or inaccurate
inference for large datasets. Further, many of these approaches have focused on
in-matrix prediction, and have not been applied to out-of-matrix predictions.

Other related literature include \citet{li2009}, where the authors proposed a
regularized matrix factorization model exploiting relation information. The
proposed model is identical to the Gaussian process factor model\footnote{See
experiments (\secref{sec:experiments}) for further discussion.} \citep{zhou12}
with an appropriate choice of kernel.
\citet{li2009b} proposed an approach for learning a kernel based on network
links that can then be applied to predictive modeling tasks. \citet{li2009c}
proposed a Bayesian probabilistic PCA model for full matrix prediction
exploiting relational data information by constructing a covariance matrix that
accounted for the relational data. An alternative approach focusing on learning
additive Gaussian process kernels was proposed by \citep{xu2009}, and an
approach for nonparametric relational data modeling using co-clustering (instead
of matrix factorization) was proposed by \citet{xu2006}. Several works have
focused on the matrix prediction task alone without the use of side information.
For example, \citet{sutskever2009} utilized the clustering of factors to model
the latent relationships as an alternative to designing covariance matrices.

\section{Proposed Approach: The Nuclear Norm Constrained
MV-GP}\label{sec:nucleargp} 

We propose nuclear norm constrained Bayesian inference for modeling low rank
transposable data as an alternative to the low rank factor approach. The proposed
approach constrains the model by directly regularizing the rank of the expected
prediction via a constraint on its nuclear norm.
Optimization with the rank constraint is computationally intractable, and the
popular factor representation results in a nonconvex optimization problem that
is susceptible to local minima \citep{dudik12}. The nuclear norm constraint has
been proposed as a tractable surrogate regularization for the low rank
constraint, which is in turn motivated by parsimony of the low rank
representation, and the superior empirical performance of low rank models in
many application domains.  The nuclear norm of a matrix variate function is
given by the sum of its singular values \citep{abernethy09}, and is the tightest
convex hull of its rank. Under certain conditions, it can be shown that nuclear
norm regularization recovers the true low rank matrix \citep{pong2010}. Further
details on the nuclear norm of matrix functions are provided in Appendix
\ref{sec:nuclearnorm}.

With no loss of generality, we assume a set of rows $\sfM$ and a set of columns
$\sfN$ of interest so $\sfL \subset \sfM \times \sfN$. Let $\bZ \in \bbR^{M
\times N}$ be the matrix of hidden variables, with $\bz = \tvec{\bZ} \in \bbR^{M
\times N}$. Given any finite index set of observations at indices $l \in \sfL$,
the finite dimensional prior distribution $P(\bz)$ is a Gaussian distribution
given by $\g{0}{\bC}$ where $\bC \in \bbR^{MN \times MN}$. We seek a
postdata distribution that optimizes \eqref{eq:conbayes} subject to
the constraint $\normt{\ev{q}{Z}} \le \eta$ where $\normt{\cdot}$ is the nuclear
norm. For any finite index set, the unconstrained Bayesian posterior
distribution is Gaussian (\secref{sec:mvgp}). Following the steps of
\secref{sec:constrain} (see also Appendix \ref{sec:con_proof}), it is straightforward to show that since the feature function
$\gamma(Z) = Z$ is linear, the constrained Bayes solution must also take a
Gaussian form. All that remains is to solve for the mean and covariance. We may
apply either the prior form \eqref{eq:conbayes} or the equivalent posterior form
\eqref{eq:conbayes_post} for constrained inference. We discuss both approaches
for illustrative purposes.

Let the Bayesian posterior be given by $
\g{\bphi}{\bSigma}$ as described in \eqref{eq:post_all} where $ \bphi =
\tvec{\bPhi} \in \bbR^{M \times N}$, and $\bSigma\in \bbR^{MN \times MN}$. Let
the postdata density be given by $\g{\bpsi}{\bS}$, where $ \bpsi =
\tvec{\bPsi} \in \bbR^{M \times N}$, and $\bS\in \bbR^{MN \times MN}$.
Using the posterior form \eqref{eq:conbayes_post}, the postdata density is
found by minimizing the KL divergence between the Gaussian distribution
$\g{\bpsi}{\bS}$ and the Bayesian posterior distribution $\g{\bphi}{\bSigma}$.
This is given by:
\begin{equation*}
\underset{\bpsi, \bS}{\min} \; \tr{\inv{\bSigma}\bS}
+ \trans{(\bphi - \bpsi)}\inv{\bSigma}(\bphi -\bpsi) - \log |\bS| +
\log|\bSigma| \;\; \sut \,
\normt{\ev{q}{Z}} \le \eta
\end{equation*}
where $\bpsi = \tvec{\bPsi}$. The optimization decouples
between the mean term $\bpsi$ and the covariance term $\bS$ as:
\begin{subequations} \label{eq:postrank}
\begin{gather}
\underset{\bpsi}{\min} \; \trans{(\bphi - \bpsi)}\inv{\bSigma}(\bphi -\bpsi)
\;\;\sut \, \normt{\ev{q}{Z}} \le \eta \label{eq:meanrank}
\\
\underset{\bS}{\min} \; \tr{\inv{\bSigma}\bS} - \log |\bS| +
\log|\bSigma| \label{eq:covrank}
\end{gather}
\end{subequations}
The minimum in terms of the covariance is achieved for $\bS = \bSigma$ and the
mean optimization is given by the solution of a constrained quadratic
optimization. 

Direct optimization of \eqref{eq:meanrank} requires the computation, storage and
inversion of the covariance matrix $\bSigma$. This may become computationally
infeasible for high dimensional data. In such situations, estimation of the
postdata mean using the prior form \eqref{eq:conbayes} is a more
computationally feasible approach. The result is the optimization problem:
\begin{equation}\label{eq:var_bnd}
\frL(\bPsi, \bS) = \underset{\bpsi, \bS}{\min} \bigg[ 
\bbE_{\bZ}[\ln p(\bZ)] - \bbE_{\bZ}[\ln p(\by,\bZ)]
\; \sut \, \normt{\ev{q}{Z}} \le \eta \bigg].
\end{equation}
Let $\bP \in \bbR^{L \times MN}$ be a selection matrix such that $\bS_{L} =
\bP\bS\trans{\bP}$ is the postdata covariance matrix of the subset of observed
entries $l \in \sfL$, and $\bC_{L} = \bP\bC\trans{\bP}$ is the prior covariance of the
corresponding subset of entries.
Evaluating expectations, the cost function \eqref{eq:var_bnd} results in the
following inference cost function (omitting terms independent of $\bpsi$ and
$\bS$):
\begin{equation*}
\frL(\bPsi, \bS) =
\underset{\{ \bpsi\, |\, \normt{\ev{q}{Z}} \le \eta\}, \bS
}{\min} \; 
\left[
\begin{array}{c}
\fracl{2\sigmad} 
\sum_{\mn \in \sfL}(y\smn-\psi\smn)^2 
+\half \trans{\bpsi}\inv{\bC}\bpsi \\
- \ln|\bS|
+\fracl{2\sigmad} \tr{ \bS_{L}} +\half \tr{\inv{\bC}\bS}  
\end{array}
\right].
\label{eq:all_cost}
\end{equation*}

First, we compute gradients with respect to $\bS$. After setting the gradients
to zero, we compute:
\begin{equation}\label{eq:qsigma}
\bS_* = \inv{\left({ \inv{\bC} + \fracl{\sigmad} \trans{\bP}\bP
}\right)} = \bC - \bC\trans{\bP}\inv{\left({\bC_L +
\fracl{\sigmad} \bI_L }\right)} \bP\bC
\end{equation}
The second equality is a consequence of the matrix inversion lemma. We note
that this is the exact same result as was found by using the posterior approach
\eqref{eq:covrank}. Next, collecting the terms involving the mean results in
the optimization problem:
\begin{equation}
\bpsi_*= \underset{\bpsi }{\arg\min} \;\fracl{2\sigmad} \sum_{m,n \in
\sfL}(y\smn-\psi\smn)^2 +\half \trans{\bpsi}\inv{\bC}\bpsi 
\quad \text{s.t.}\; \normt{\ev{q}{Z}} \le \eta \label{eq:qmean}
\end{equation}
This is a convex regularized least squares problem with a convex
constraint set. Hence, \eqref{eq:qmean} is convex, and $\bpsi_*$ is unique.
Using the Kronecker identity, we can re-write the cost function in parameter
matrix form. We can also replace the nuclear norm constraint with the equivalent
regularizer weighed by $\lambda$. This leads to
the equivalent optimization problem:
\begin{equation}\label{eq:qmean_reg}
\bPsi_*
=\underset{\Psi }{\arg \min} \; \fracl{2\sigmad} \sum_{m,n \in
\sfL}(y\smn-\psi \smn)^2
+\half \tr{\trans{\bPsi}\inv{\km}\bPsi\inv{\kn}}
+\lambda \normt{\ev{q}{Z}}.
\end{equation}

The final step is to define the term $\normt{\ev{q}{Z}}$. We note that since the
prior distribution is a Gaussian process, a valid postdata distribution must
extend to arbitrary index sets. Hence the postdata mean is a matrix-variate
function. The parametric representation of the postdata mean can be defined
using the posterior distribution of the Gaussian process outlined by
\citet{csato02} and applying the representation theorem \eqref{eq:representer}.
Thus, we recover the parametric form of the mean function as $\bPsi = \km
\bAlpha \kn$ where $\bAlpha \in \bbR^{M \times N}$. We may now solve for
$\bAlpha$ directly:
\begin{equation}\label{eq:qmean_kernel}
\bAlpha_*
=\underset{ \bAlpha }{\arg \min} \; \frac{1}{2\sigmad} \sum_{\mn \in
\sfL} \left(y\smn-(\km\bAlpha\kn)\smn \right)^2
+\half \tr{\trans{\bAlpha}\km\bAlpha \kn}
+\lambda \normht{\psi_\bAlpha}{\hk}.
\end{equation}
where $\psi_\bAlpha$ is the mean function corresponding to
the parameter $\bAlpha$ (see \eqref{eq:representer}), and $\normht{\cdot}{\hk}$
represents the nuclear norm in the Hilbert space $\hk$ (defined in Appendix
\ref{sec:nuclearnorm}). 
We also note that the optimization problem \eqref{eq:qmean_kernel} is strongly
convex.

We now seek to extend the solution from the finite observed index set to the
nonparametric domain. Our approach will rely on Kolmogorov's Extension theorem
\citep{bauer96} which provides a mechanism for describing infinite dimensional
random processes via their finite dimensional marginals \citep{orbanz10}.
We will apply the theorem to extend the solution estimated by \eqref{eq:var_bnd}
using a finite index to a corresponding nonparametric Gaussian process. This
will be achieved by showing that the solution can be extended to an arbitrary
index set with a consistent functional form for the mean and the covariance.

\begin{theorem}\label{thrm:gp}
The postdata distribution $\g{\bpsi}{\bS}$ is a finite
dimensional representation of the Gaussian process $\gp{\psi}{S}$ sampled
at indices $\sfL$ where the mean function $\psi$ is given by
\eqref{eq:qmean_kernel} and the covariance function $S$ is given by
\eqref{eq:post_var}.
\end{theorem}
\emph{Sketch of proof:}
The requirements of Kolmogorov's extension theorem can be reduced to a proof
that for a fixed training set $\data$, the postdata distribution of the
superset $(\sfM \times \sfN)\cup (m',n')$ has a consistent function
representation\footnote{See \citep[Section 2.2]{rasmussen05} for an analogous
proof applied to Gaussian process regression.}.
The mean and covariance of the postdata density are decoupled in the
optimization and the postdata covariance function can be computed in closed
form. Thus, for the covariance, this follows trivially from the functional form
of \eqref{eq:post_var}. The functional form of the mean follows from the finite
representation \eqref{eq:representer} that solves the optimization problem
\eqref{eq:qmean_kernel}. Note that the solution does not change with the
addition of indices $l'=(m',n') \notin \sfL$ without corresponding
observations $y_{l'}$. Uniqueness of the solution follows from the strong
convexity of \eqref{eq:qmean_kernel}.
We refer the reader to the dissertation \citep{koyejo2013z} for
further details.

\subsection{Alternative Representation of the Nuclear Norm Constrained Inference}\label{sec:nuclear_inference}

The mean function optimization \eqref{eq:qmean_kernel} may also be represented
in terms of matrix parameters that are amenable to direct optimization. With the
index set fixed, compute a basis. $\gm\in \bbR^{M \times D\sm}$ and $\gn\in
\bbR^{N \times D\sn}$ such that $\km = \gm\trans{\gm}$ and $\kn =
\gn\trans{\gn}$.
The mean function can be re-parameterized as $\psi(m,n) =
\gm(m)\bB\trans{\gn(n)}$, where $\bB \in \bbR^{D\sm \times D\sn}$.
The nuclear norm of $\psi$ can now be computed directly as the nuclear norm of
the parameter matrix \citep[Theorem 3]{abernethy09}.
The resulting optimization problem is:
\begin{equation}\label{eq:qmean_matrix}
\bB_*=\underset{ \bB }{\arg \min} \;
\frac{1}{2\sigmad} 
\sum_{\mn \in
\sfL}\left(y\smn-(\gm\bB\trans{\gn})\smn \right)^2
+\half \normfs{\bB}+ \lambda \normt{\bB}.
\end{equation}
where $\bB$ is the estimated parameter matrix, and $\normfs{\cdot}$ and 
$\normt{\cdot}$ represent the matrix squared Frobenius norm and the matrix
nuclear norm respectively. In this form, the mean function can be
estimated directly using standard solvers for large scale nuclear norm
constrained optimization (e.g. \citet{dudik12, laue12}). 

To improve scalability, large scale nuclear norm regularized solvers generally
represent the parameter matrix in low rank form, avoiding storage of the full
matrix.
Further, the rank of the parameter matrix is automatically estimated during the
optimization. We provide a short summary of the approaches in \citet{dudik12}
and \citet{laue12}. Interested readers are referred to the relevant papers for
further details. The parameter matrix can be estimated starting from a rank one
solution, then the rank is increased until additional factors do not improve the
cost any further. The first step consists of determining a good descent
direction, and the second step consists of optimizing the factors given the
initial direction. In the first step, a descent direction is determined by
computing (approximate) singular singular vectors associated with the maximum
signular value of a sparse gradient matrix. This step does not need to be
accurate and is usually achieved using a few iterations of the power method. The
factor optimization in the second step is analogous to the standard matrix
factorization optimization, so the large scale nuclear norm solvers mainly
differ from standard matrix factorization in the determination of an initial
descent direction\footnote{In contrast, matrix factorization is generally
randomly initialized.}, and in the automatic determination of the number of
required factors i.e. the rank.
Thus the computational requirements of large scale nuclear norm regularized
regression are comparable to standard matrix factorization methods.

\section{Experiments}\label{sec:experiments}

We completed experiments with transposable datasets from the disease-gene
association domain and the recommender system domain. {\bf Prior covariances:}
All the datasets studied consist of transposable data matrices with
corresponding row and/or column graphs. We experimented with the identity prior
covariance $\bC = \bI$, where $\bI$ is the identity matrix, and the diffusion
prior covariance \citep{smola03} given as $\bC = \exp{(-a\bL)} + b\bI$, where
$\bL$ is the normalized graph Laplacian matrix. Let $\bA$ be the adjacency
matrix for the graph and $\bD$ be a diagonal matrix with entries $\bD_{i,i}=
(\bA\mathbf{1})_i$. The normalized Laplacian matrix is computed as $\bL = \bI -
\bD^{-\half}\bA\bD^{-\half}$. We set the $a=b=1$. No further optimization was
performed, and more detailed experimental validation of covariance parameter
selection is left for future work.

{\bf Models:} We present results for the proposed constrained MV-GP approach
({\bf Con. MV-GP}), and the special cases using only the nuclear
norm\footnote{The nuclear norm is also known as the trace norm.} ({\bf
Trace GP}) and using only
the Hilbert norm ({\bf MV-GP}) i.e.
the standard MV-GP regression. To the best of our knowledge, the special case of
Trace GP is a novel contribution. As baselines, we implemented kernelized
probabilistic matrix factorization ({\bf KPMF}) \citep{zhou12} and probabilistic
matrix factorization ({\bf PMF}) \citep{ruslan08} using rank 5 and rank 20
factors.
PMF is identical to KPMF using an identity covariance. KPMF has been shown to
outperform PMF and and other baseline models in various domains. We note that
the rank constraint ensures that all of the proposed models except for MV-GP can
be used for in-matrix predictions even with the identity prior covariance.
Out-of-matrix predictions require the use of other covariance matrices.

We implemented Con. MV-GP using the representation outlined in
\secref{sec:nuclear_inference}. The Cholesky decomposition of the covariance
matrices was used as the basis representation. The model hyperparameter
$\lambda$ was selected using $5$ values logarithmically spaced between $10^{-3}$
and $10^{3}$ and the noise hyperparameter was selected $\sigma^2$ using 20
values logarithmically spaced between $10^{-3}$ and $10^{3}$ for all the models.
We experimented with learning the data noise variance term $\sigma^2$, but found
the results worse than using parameter selection. In particular, the estimated
noise variance often approached zero - indicating overfitting. A possible
solution we plan to explore is to introduce a prior distribution for $\sigma^2$
( see e.g. Bayesian linear regression in \citet[Chapter 3.3]{bishop06}) that may
help to regularize the noise term away from zero.

The standard MV-GP is often implemented as a scalar GP with the row and column
prior covariance matrices multiplied as shown in \eqref{eq:post_all}.
We found this ``direct" approach computationally intractable as the memory
requirements scale quadratically with the size of the observed transposable data
matrix. Instead, we implemented the MV-GP in matrix form as a special case of
\eqref{eq:qmean_matrix} with $\lambda = 0$. This allowed us to scale the model
to the larger datasets at the expense of more computation. The nuclear norm
regularized optimization in \eqref{eq:qmean_matrix} was solved using the large
scale approach of \citet{laue12}. All numerical optimization was implemented
using the limited memory Broyden-Fletcher-Goldfarb-Shanno (L-BFGS) algorithm.

{\bf Experiment design and cross validation:} We performed two kinds of
experiments. In the rest of this discussion, ``rows'' will refer to either the
disease (disease-gene prediction) or the user (recommender system). The
\emph{known rows} experiment was designed to evaluate the performance of the
model for entries selected randomly over the observed values in the matrix.
In contrast, the \emph{new rows} experiment was designed to evaluate the
generalization ability of the model for new rows not observed in the training
set. We partitioned each dataset into five-fold crossvalidation sets. The model
was trained on $4$ of the $5$ sets and tested on the held out set. The results
presented are the averaged $5$-fold cross validation performance. For the
``known row'' experiments, the cross validation sets were randomly selected over
the matrix. For the \emph{new row} experiments, the cross validation was
performed row-wise, i.e., we selected training set row and test set rows. Note
that the identity prior covariance cannot be used for new row prediction, but
due to the low rank constraint, it can be used for known row prediction.

\subsection{Disease-Gene Prediction}\label{sec:disease-gene}

Genes are segments of DNA that determine specific characteristics; over 20,000
genes have been identified in humans, which interact to regulate various
functions in the body. Researchers have identified thousands of diseases,
including various cancers and respiratory diseases such as asthma
\citep{gene11}, caused by mutations in these genes. Genetic association studies
\citep{mccarthy08} are the standard approach for discovering disease-causing
genes. However, these studies are often tedious and expensive to conduct. Hence,
computational methods that can reduce the search space by predicting the list of
candidate genes associated with a given disease are of significant scientific
interest.
The disease gene prediction task has been the subject of a significant amount of
study in recent years \citep{vanunu10, li10, mordelet11, singh-blom13}. The task
is challenging because all the observed responses correspond to known
associations, and there are no reliable negative examples. Disease gene
association shares the binary matrix representation of the \emph{one class}
(also known as implicit feedback) matrix prediction studied in the collaborative
filtering literature \citep{pan08, hu08}.

{\bf Additional baseline: } In addition to the matrix factorization baseline
models, we compared our proposed approach to ProDiGe \citep{mordelet11}; a
start-of-the-art approach that has been shown to be superior to previous
top-performing approaches, including distance-based learning methods like
Endeavour \citep{aerts06} and label propagation methods like PRINCE
\citep{vanunu10}. ProDiGe estimates the prioritization function using a
multitask support vector machine (SVM) trained with the gene prior covariance
and disease prior covariance as kernels. Of the models implemented, ProDiGe is
most similar to the MV-GP. In fact, MV-GP and ProDiGe mainly differ in their
loss functions (squared loss and hinge loss respectively). The SVM regularization
parameter for ProDiGe was selected from $\{10^{-3}, 10^{-2},\ldots , 10^3\}$.
We note that also PMF represents the matrix factorization baseline often applied
to similar implicit feedback datasets in the recommendation system literature
\citep{pan08}.

{\bf Sampling ``negative" entries:} Following \citet{mordelet11}, we sampled the
unknown entries as ``negative'' observations randomly over the disease-gene
association matrix. We sampled $5$ different negatively labeled item sets. All
models were trained with the positive set combined with one of the negative
labeled sets. The model scores were computed by averaging the scores over the 5
trained models. All models were trained using the same samples.

{\bf Metrics: } Experimental validation of disease-gene associations in a
laboratory can be time consuming and costly, so only a small set of the top
ranked predictions are of practical interest. Hence, we focus on metrics that
capture the ranking behavior of the model at the top of the ranked list. All the
ranking metrics were computed on the test set after removing \emph{all} genes
that had been observed in the training set. We computed precision
($\pres{k}$) and recall ($\recall{k}$) where $k = 1 \ldots 20$. 
Let $\bg_l$ denote the labels of gene $l$ as sorted by the predicted scores of
the trained regression model, and let $G_m = \sum_{l} \indicator{\bg_l=1}$ be
the total number of relevant genes for disease $m$ in the test data  after
removing relevant genes observed in the training data.
The precision at $k$ computes the fraction of relevant genes retrieved out off
all retrieved genes at position $k$.  The recall at $k$ computes the fraction of
relevant genes retrieved out of all relevant genes that can be retrieved with a
list of length $k$. These are computed as:
\begin{equation*} 
  	\pres{k} = \frac{\sum_{l=1}^k \indicator{\bg_l=1} }{k}, \quad
	\recall{k} = \frac{\sum_{l=1}^k \indicator{\bg_l=1} }  
  	{G_m}.
\end{equation*}
All metrics were computed per disease and then averaged over all the diseases in
the test set. Model selection was computed separately per metric. Higher values
reflect better performance for the $\pres{k}$ and $\recall{k}$ metrics and their
maximum value is $1.0$.

{\bf Datasets: } We trained and evaluated our models using two sets of
gene-disease association data curated from the literature. The first, which we
call the \textbf{OMIM data set}, is based on the Online Mendelian Inheritance in
Man (OMIM) database and is representative of the candidate gene prediction task
for monogenic or near monogenic diseases, i.e., diseases caused by only one or
at most a few genes.
The data matrix contains a total of $M = \text{3,210}$ diseases, $N =
\text{13,614}$ genes, and $T = \text{3,636}$ known associations (data density of
0.0083\%). We note that the extreme sparsity of this data set makes the
prediction problem extremely difficult.
The second dataset, which we call the \textbf{Medline data set}, is a much
larger data set and is representative of predicting candidate genes for both
monogenic as well as polygenic diseases, i.e., diseases caused by the
interactions of tens or even hundreds of genes. The set of genes in this data
set is defined using the NCBI ENTREZ Gene database \citep{maglott11}, and the
set of diseases is defined using the ``Disease'' branch of the NIH Medical
Subject Headings (MeSH) ontology \citep{mesh}. We extracted co-citations of
these genes and diseases from the PubMed/Medline database \citep{pubmed} to
identify positive gene-disease associations. This resulting data set contains a
total of of $M = \text{4,496}$ diseases, $N = \text{21,243}$ genes, and $T =
\text{250,190}$ known associations (data density of 0.36\%).

Information about biological interactions among genes and known relationships
among diseases were used to improve the accuracy of our model, since similar
diseases very often have similar genetic causes. We derive \textbf{gene
networks} from the HumanNet database \citep{lee11}, a genome-wide functional
network of human genes constructed using multiple lines of evidence, including
gene co-expression, protein-protein interaction data, and networks from other
species. For both the OMIM and Medline data sets, our gene-gene interaction
network contains a total of $\text{433,224}$ links. Our \textbf{disease network}
is derived from the term hierarchy established in the 2011 release of the MeSH
ontology. The disease network for the Medline data set contains a total of
$\text{13,922}$ links. However, because we do not have a direct mapping of OMIM
diseases to MeSH terms, we do not use a disease network for the OMIM data set.
As a result, we are unable to test our model's ability to produce predictions
for ``new'' diseases, i.e., diseases with no associated genes in the
training set.

\begin{table}[!th]
\centering
\caption{OMIM disease-gene dataset. Avg. (std.) $\pres{20}$ and
$\recall{20}$ performance. (I): Identity prior covariance
(D): Diffusion prior covariance. Low precision values in OMIM data are due to the high
class imbalance of the test data (average of 1.2 genes per disease).
\label{tab:marcotte_mat_all}}
\begin{tabular}{lll}
\hline\noalign{\smallskip}
Model&$\pres{20}$&$\recall{20}$\\
\noalign{\smallskip}\hline\noalign{\smallskip}
MV-GP (D) & {0.000 (0.000)} & {0.003 (0.002)} \\
\hline
Con. MV-GP (D) & {0.008 (0.001)} & {0.146 (0.031)} \\
\hline
Con. MV-GP (I) & {\bf 0.010 (0.001)} & {\bf 0.175 (0.025)} \\
\hline
Trace GP (D) & {0.006 (0.001)} & {0.117 (0.021)} \\
\hline
Trace GP (I) & {0.009 (0.001)} & {0.157 (0.023)} \\
\hline\hline\noalign{\smallskip}
KPMF-5 (D) & {\bf 0.010 (0.001)} & {0.167 (0.028)} \\
\hline
PMF-5 (I) & {0.002 (0.000)} & {0.034 (0.004)} \\
\hline
KPMF-20 (D) & {0.009 (0.002)} & {0.161 (0.040)} \\
\hline
PMF-20 (I) & {0.002 (0.000)} & {0.039 (0.008)} \\
\hline
ProDiGe & {0.000 (0.000)} & {0.001 (0.003)} \\
\noalign{\smallskip}\hline
\end{tabular}
\end{table}

\begin{table}[!th] 
\centering
\caption{ Medline disease-gene dataset. Avg. (std.)
$\pres{20}$ and $\recall{20}$ performance. (I): Identity prior covariance (D):
Diffusion prior covariance. The dataset contains an average of 59.2 test genes
per disease. The identity prior covariance does not generalize to new diseases.
ProDiGe was unable to scale to the full curated dataset.}
\label{tab:our_comb_all}
\begin{tabular}{lllll}
\hline\noalign{\smallskip}
&\multicolumn{2}{l}{Known diseases}&\multicolumn{2}{l}{New Diseases}\\
\noalign{\smallskip}\hline\noalign{\smallskip}
Model&$\pres{20}$&$\recall{20}$&$\pres{20}$&$\recall{20}$\\
\noalign{\smallskip}\hline\noalign{\smallskip}
MV-GP (D) & {0.022 (0.000)} & {0.049 (0.002)} 
& {0.069 (0.020)} & {0.091 (0.022)}\\
\hline
Con. MV-GP (D) & {0.078 (0.001)} & {0.131 (0.004)} 
& {\bf 0.137 (0.029)} & {\bf 0.181 (0.026)} \\
\hline
Con. MV-GP (I) & {\bf 0.126 (0.001)} & {\bf 0.216 (0.002)} 
& -- & --\\
\hline
Trace GP (D) & {0.078 (0.001)} & {0.131 (0.004)} 
& {\bf 0.137 (0.029)} & {\bf 0.181 (0.026)} \\
\hline
Trace GP (I) & {0.091 (0.001)} & {0.152 (0.004)}
& -- & --\\
\hline\hline\noalign{\smallskip}
KPMF-5 (D) & {0.085 (0.001)} & {0.142 (0.004)} 
& {0.136 (0.032)} & {0.179 (0.032)} \\
\hline
PMF-5 (I) & {0.079 (0.002)} & {0.133 (0.003)}
& -- & --\\
\hline
KPMF-20 (D) & {0.091 (0.001)} & {0.151 (0.004)} 
& {0.136 (0.032)} & {0.179 (0.032)} \\
\hline
PMF-20 (I) & {0.078 (0.001)} & {0.131 (0.002)} 
& -- & --\\
\noalign{\smallskip}\hline
\end{tabular}
\end{table}

\begin{figure}[!th]
    \begin{center}
    \subfigure[OMIM precision$@k$]{
    \includegraphics[width=0.45\textwidth]{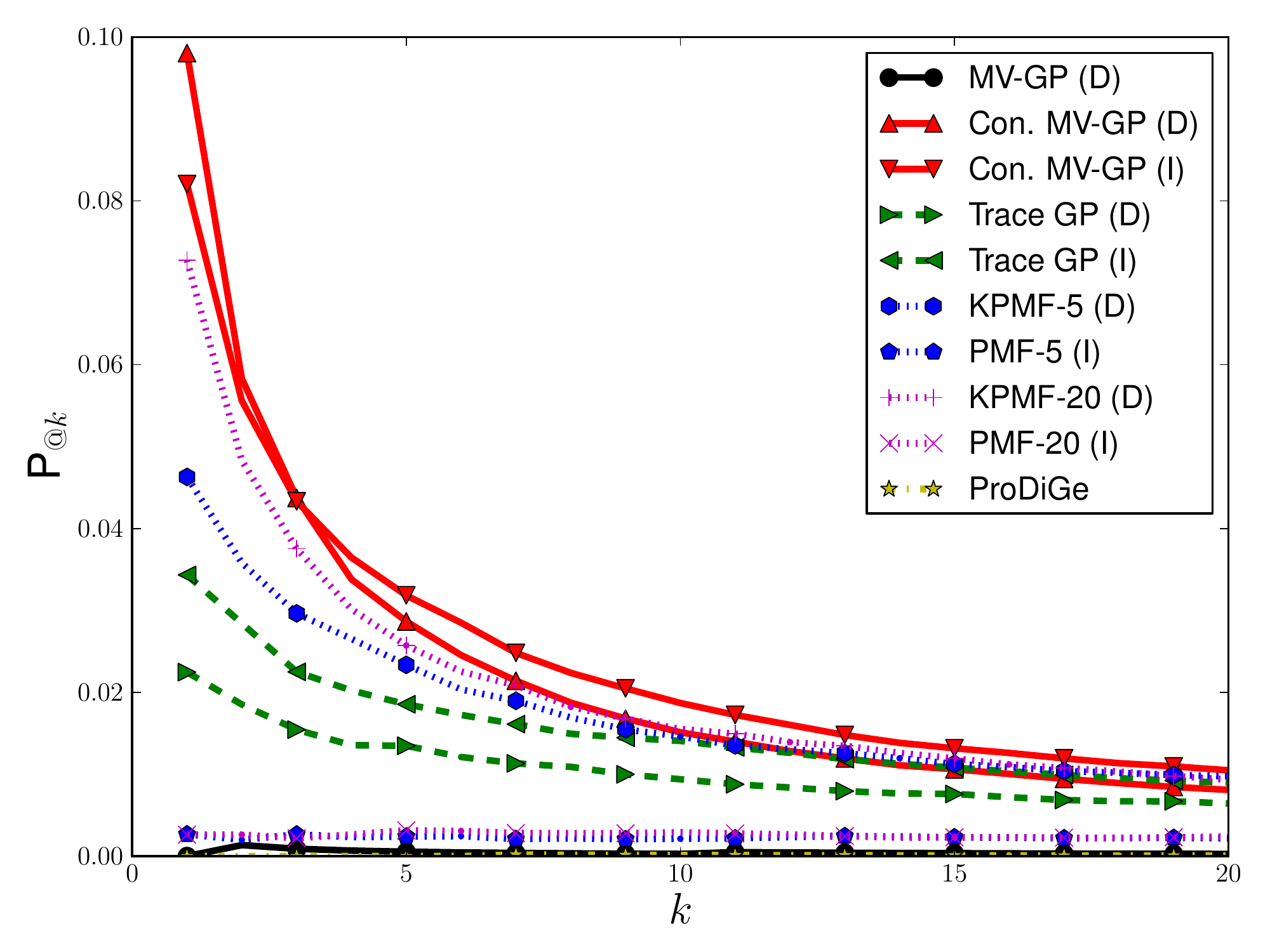}
    \label{fig:marcotte_mat_all_pre}
	}
    \quad
    \subfigure[OMIM recall$@k$]{
    \includegraphics[width=0.45\textwidth]{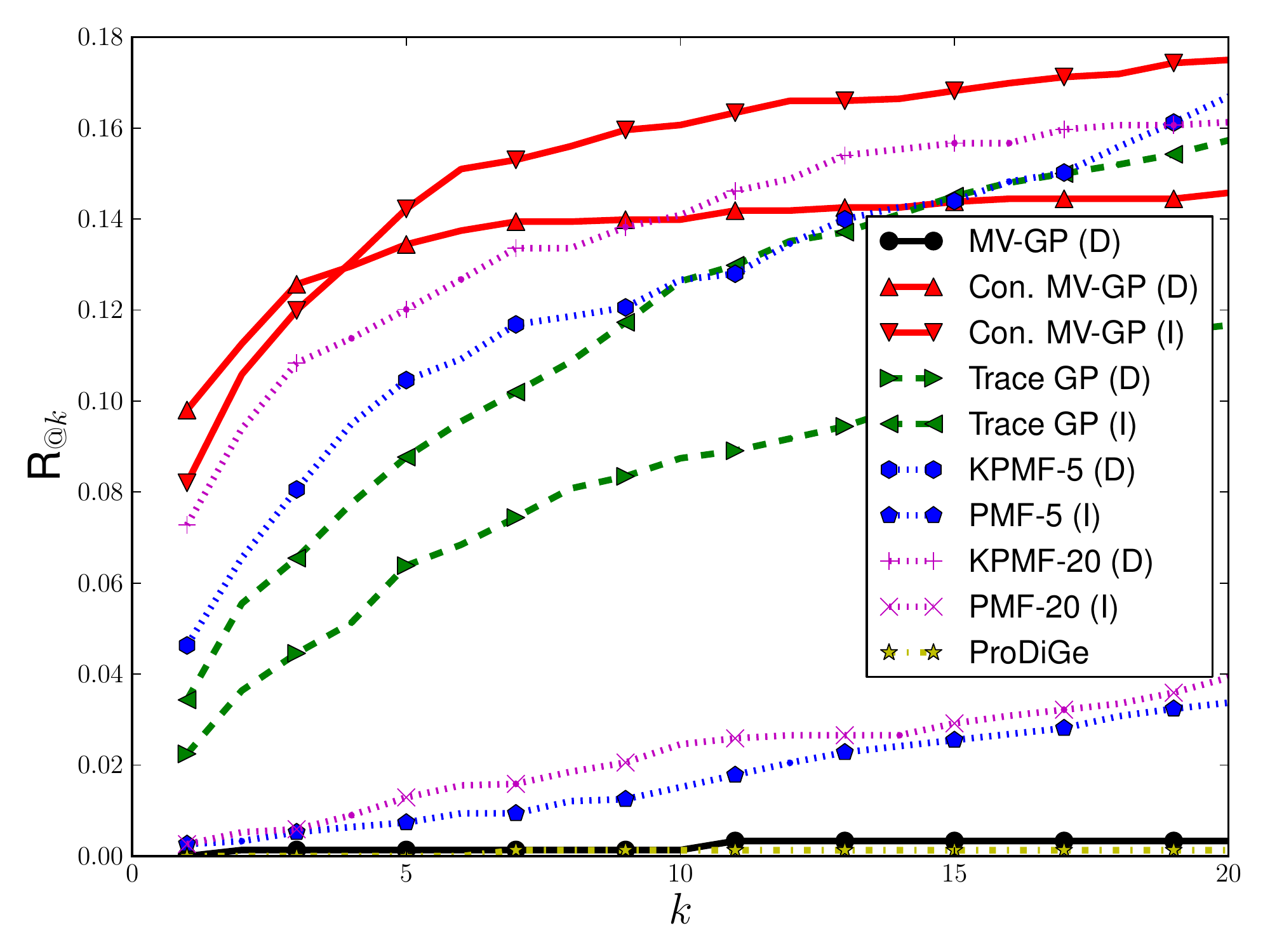}
    \label{fig:marcotte_mat_all_rec}
	}%

    \subfigure[Medline (known diseases) precision$@k$]{
    \includegraphics[width=0.45\textwidth]{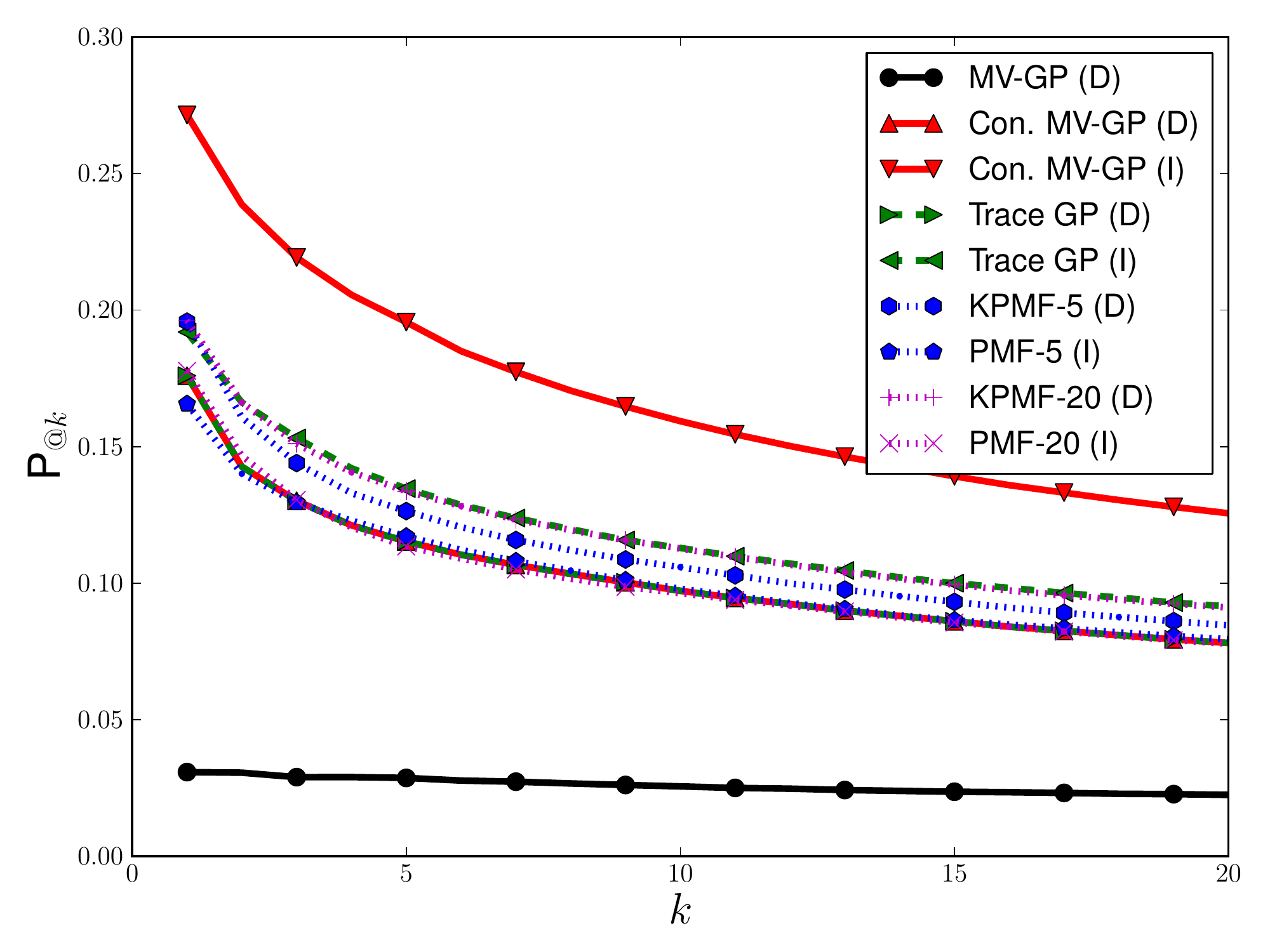}
    \label{fig:our_mat_all_pre}
	}
	\quad
	\subfigure[Medline (known diseases) recall$@k$]{
    \includegraphics[width=0.45\textwidth]{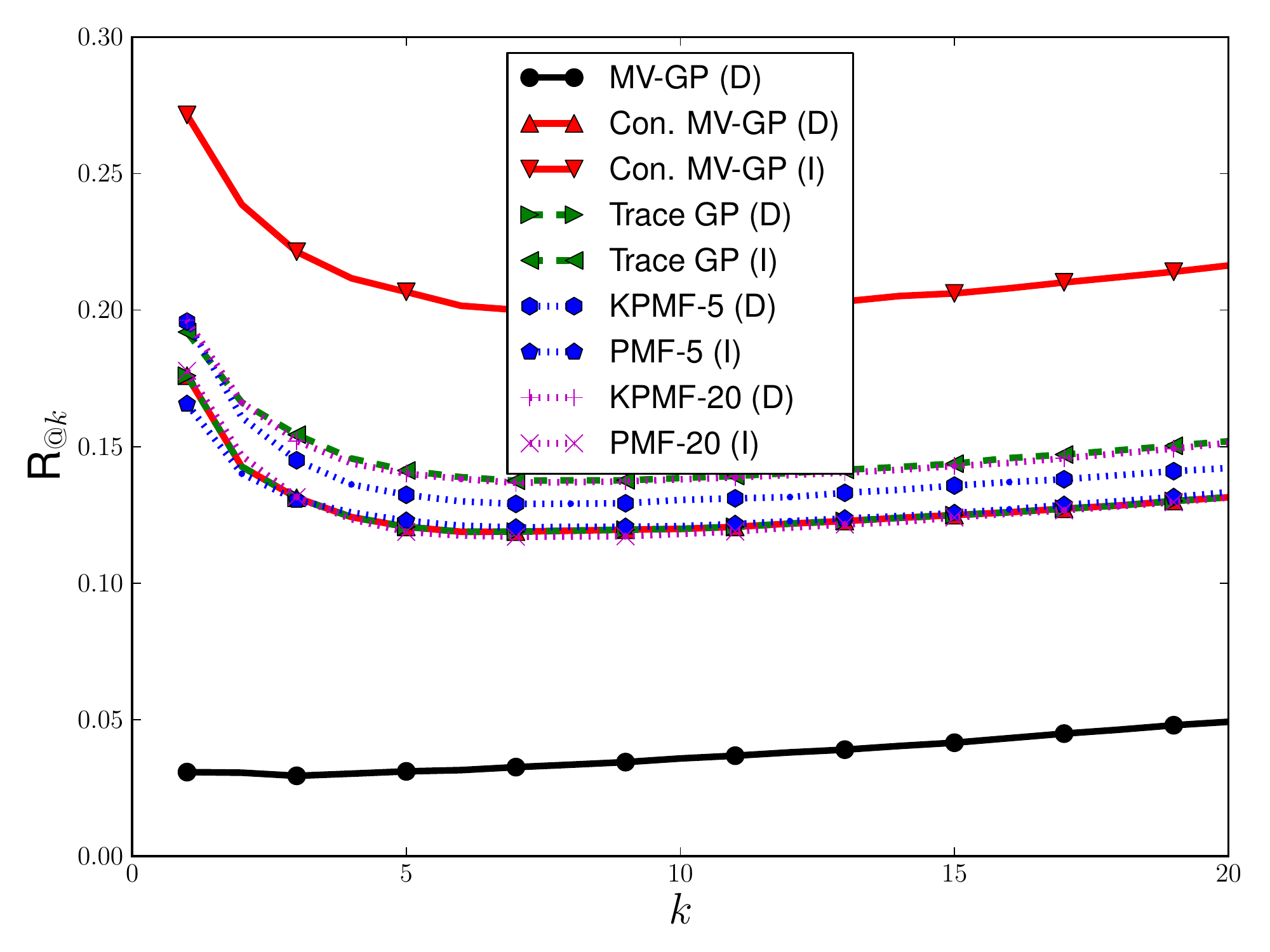}
    \label{fig:our_mat_all_rec}
	}

    \subfigure[Medline (new diseases) precision$@k$]{
    \includegraphics[width=0.45\textwidth]{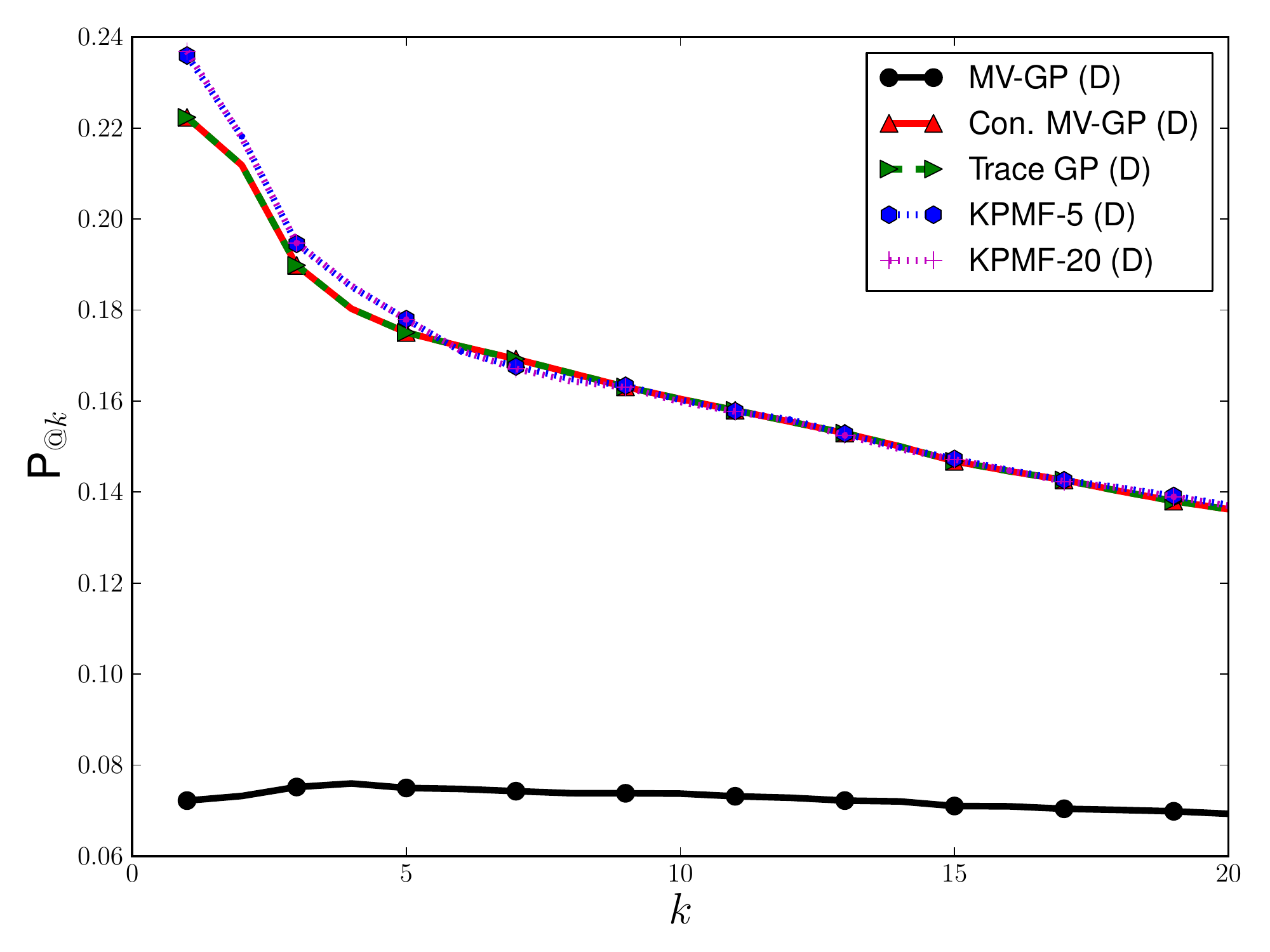}
    \label{fig:our_cold_all_pre}
}
    \quad \subfigure[Medline (new diseases) recall$@k$]{
    \includegraphics[width=0.45\textwidth]{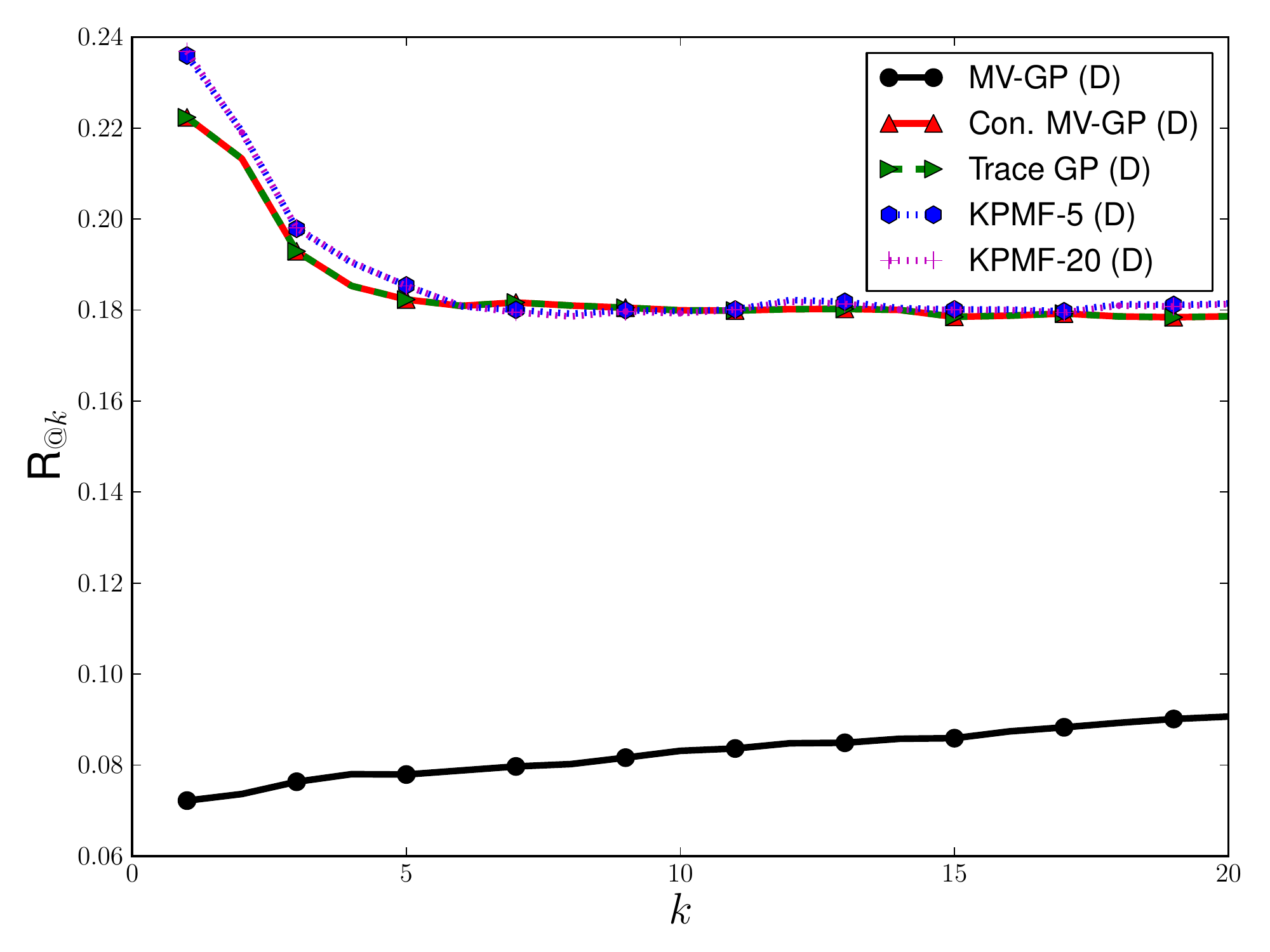}
    \label{fig:our_cold_all_rec}
}
	\end{center}
    \caption{Disease-gene prediction. Precision (left) and Recall (right) $@k=1,
    2, \ldots, 20$.
    (I): Identity prior covariance, (D): Diffusion prior covariance. Low
    precision values in OMIM are due to the high class imbalance in the test
    data (avg of 1.2 genes per disease). The identity prior covariance does not
    generalize to new diseases. Constrained MV-GP out-performs ProDiGe (domain
    specific baseline), KPMF and PMF.  ProDiGe was unable to scale to the full
    curated dataset (see text).}\label{fig:gene_expts}
\end{figure}

The OMIM dataset contains an average of $1.2$ test genes (positive items) per
disease, and the model is required to rank more than $13,000$ genes per disease.
Hence, the gene prediction task is particularly challenging. This difficulty is
reflected in the low precision values observed in \tabref{tab:marcotte_mat_all}
and \figref{fig:marcotte_mat_all_pre}. Despite this extreme sparsity, we found
that the proposed approaches (Con. MV-GP and Trace GP) performed as well or
better than the matrix factorization baselines (KMPF, PMF), and significantly
outperformed the domain specific baseline (ProdiGe). In fact, both full rank
models (MV-GP and ProDiGe) performed poorly, suggesting the importance of the
low rank / nuclear norm constraint. The results in
\figref{fig:marcotte_mat_all_pre} and \figref{fig:marcotte_mat_all_rec} further
highlight the performance of the proposed models at the very top of the list.

We were unable to run ProDiGe on the Medline dataset due computational issues.
In particular, the implementation of ProDiGe requires the full kernel matrix as
an input. The memory required to store the full kernel is quadratic in the
transposable data size. We did not pursue an alternative implementation with
reduced memory requirements as experiments with OMIM and initial experiments
with subsampled data indicated inferior performance. The Medline dataset
contained an average of $59.2$ positive items per disease. Correspondingly, the
tested models achieved a higher precision than in the OMIM dataset. Our
experimental results (\tabref{tab:our_comb_all}) show that the
proposed models (Con. MV-GP, Trace GP) significantly outperformed the matrix
factorization baselines (PMF, KMPF) on the known diseases, and performed as
least as well as KMPF on the new diseases. The results in
\figref{fig:our_mat_all_pre} and \figref{fig:our_mat_all_rec} show that the
proposed models outperform the baselines for known diseases prediction at all
levels of precision and recall we measured. The results for new disease
prediction in \figref{fig:our_cold_all_pre} and \figref{fig:our_cold_all_rec}
show similar performance for both approaches on the new diseases.

In summary, the presented results suggest that the low rank constraint is useful
for describing the structure of disease-gene association. We also found that in
all the datasets, the constrained Bayesian models (Con. MV-GP and Trace GP)
performed the same or better than the Bayesian factor models (KPMF and PMF) and
the unconstrained Bayesian model (MV-GP). This shows the utility of the
constrained Bayesian inference approach as compared to the Bayesian factor model
approach. Constrained MV-GP with the identity kernel was the best single
performing method, matching results in the literature suggesting that the
network information is not always helpful for in-matrix predictions
\citep{koyejo11, zhou12}, though it remains essential for generalization beyond
the training matrix. Future work will include further examination of these
issues.

\subsection{Recommender Systems}\label{sec:recsys}

The goal of a recommender system is to suggest items to users based on past
feedback and other user and item information. Recommender systems may also be
used for targeted advertising and other personalized services. The low rank
matrix factorization approach has proven to be a popular and effective model for
the recommender systems data \citep{ruslan08, koren09, koyejo11}. Several
authors \citep{yu07, zhu09, zhou12} have studied the factor GP approach for
recommender systems, and have shown that prior covariances extracted from the
social network can improve the prediction accuracy and may be used to provide
predictions with no training ratings \citep{koyejo11, zhou12}. Kernelized
probabilistic matrix factorization (KPMF) is of particular interest, as it has
been shown to outperform PMF \citep{ruslan08} and SoRec \citep{ma08}, strong
baseline methods for predicting user item preferences with social network side
information.

{\bf Metrics} The model performance was measured using a combination of
regression and ranking metrics.
Recommender systems are typically most concerned with presenting the few items
that the user is very likely to be interested in, and accurately predicting the
score of the other items is less important. Several authors \citep{steck10a,
steck10b} have shown that measuring the recall ($\recall{k}$) of the top
relevant items compared to all available items can provide an unbiased estimate
of the predicted ranking.
As suggested by \citep{steck10a} we measure the ability of the model
to predict relevant items (ratings greater than 4)  ahead of other entries (both
missing and observed entries with rating less than or equal to 4) using recall
at 20 ($\recall{20}$). Recall per user was computed on the test set after
removing \emph{all} items that had been observed in the training set, and
averaged over all users. For regression, we used the root mean square error
($\rmse$) metric \citep{koren09} given by $\sqrt{\frac{1}{L}\sum_{l=1}^L (y_l -
\hat{y}_l)^2}$ where $\hat{y}_l$ is the prediction for index $l$. Lower values
reflect better performance for the $\rmse$. 

{\bf Datasets: } We trained and evaluated our models using two publicly
available recommender systems datasets with social network side information -
Flixster and Epinions datasets. \textbf{Flixster}\footnote{www.flixster.com} is
a website where users share film reviews and ratings. The users can also signify
social connections. We utilized the dataset described by \citep{jamali10} which
contains a ratings matrix and the social network. We selected the $M=5,000$
users with the most friends in the network and $N=5,000$ movies with the most
ratings. This resulted in a matrix with $L=33,182$ (density $=0.001\%$) ratings
and $211,702$ undirected user social connections. The identity prior covariance
was used for the movies.
Ratings in Flixster take one of 10 values in the set $\{0.5, 1, 1.5, \ldots ,
5.0 \}$. \textbf{Epinions}\footnote{www.epinions.com} is an item review site
where users can also specify directed association by signifying a trust link. We
utilized the extended Epinions dataset \citep{paolo06} and converted all the
directed trust links into undirected links. We selected the $M=5,000$ users with
the most trust links in the network and $N=5,000$ movies with the most ratings.
This resulted in a matrix with $L=187,163$ (density $0.007\%$) ratings and
$550,298$ user social connections. The identity prior covariance was used for
the items. Ratings in the Epinions dataset take one of five values in the set
$\{1.0, 2.0, \ldots, 5.0 \}$.

\begin{table}[!th] 
\centering
\caption{Flixster dataset. Avg.  (std.) $\rmse$ and
$\recall{20}$ performance comparison. Smaller $\rmse$ indicates better
performance, Larger $\recall{20}$ indicates better performance.  (I): Identity
prior covariance,  (D):
Diffusion prior covariance.}
\label{tab:flixtr_mat_comb}
\begin{tabular}{lllll}
\hline\noalign{\smallskip}
&\multicolumn{2}{l}{Known Users}&\multicolumn{2}{l}{New Users}\\
\noalign{\smallskip}\hline\noalign{\smallskip}
Model&\rmse & $\recall{20}$ &\rmse & $\recall{20}$\\
\noalign{\smallskip}\hline\noalign{\smallskip}
MV-GP (D) & {1.066 (0.006)} & {0.067 (0.008)}
& {1.066 (0.088)} & {\bf 0.075 (0.017)} \\
\hline
Con. MV-GP (D) & {0.989 (0.002)} & {0.092 (0.012)}
 & {1.066 (0.088)} & {\bf 0.075 (0.017)} \\
\hline
Con. MV-GP (I) & {\bf 0.982 (0.001)} & {\bf 0.104 (0.004)}
& -- & -- \\
\hline
Trace GP (D) & {0.989 (0.002)} & {0.088 (0.008)} 
& {1.066 (0.088)} & {0.069 (0.015)} \\
\hline
Trace GP (I) & {\bf 0.982 (0.001)} & {0.093 (0.003)}
& -- & -- \\
\hline\hline\noalign{\smallskip}
KPMF-5 (D) & {0.993 (0.003)} & {0.064 (0.012)}
& {1.066 (0.088)} & {0.062 (0.014)} \\
\hline
PMF-5 (I) & {0.995 (0.003)} & {0.052 (0.006)}
& -- & -- \\
\hline
KPMF-20 (D) & {0.986 (0.001)} & {0.069 (0.007)}
& {1.066 (0.088)} & {0.069 (0.015)} \\
\hline
PMF-20 (I) & {0.989 (0.002)} & {0.070 (0.003)}
& -- & -- \\
\noalign{\smallskip}\hline
\end{tabular}
\end{table}

\begin{table}[!th] 
\centering
\caption{Epinions dataset. Avg.  (std.) $\rmse$ and
$\recall{20}$ performance comparison. Smaller $\rmse$ indicates better
performance, Larger $\recall{20}$ indicates better performance.  (I): Identity
prior covariance,  (D): Diffusion prior covariance. }
\label{tab:epinions_mat_comb}
\begin{tabular}{lllll}
\hline\noalign{\smallskip}
&\multicolumn{2}{l}{Known Users}&\multicolumn{2}{l}{New Users}\\
\noalign{\smallskip}\hline\noalign{\smallskip}
Model&\rmse&$\recall{20}$ &\rmse&$\recall{20}$\\
\noalign{\smallskip}\hline\noalign{\smallskip}
MV-GP (D) & {0.323 (0.007)} & {0.016 (0.000)} 
& {0.329 (0.020)} & {0.029 (0.002)} \\
\hline
Con. MV-GP (D) & {0.273 (0.005)} & {0.023 (0.001)} 
& {0.307 (0.022)} & {\bf 0.036 (0.009)} \\
\hline
Con. MV-GP (I) & {0.274 (0.006)} & {\bf 0.046 (0.002)} 
& -- & --\\
\hline
Trace GP (D) & {0.273 (0.005)} & {0.022 (0.001)} 
& {0.307 (0.022)} & {0.035 (0.009)} \\
\hline
Trace GP (I) & {0.274 (0.006)} & {0.041 (0.003)}
& -- & --\\
\hline\hline\noalign{\smallskip}
KPMF-5 (D) & {0.274 (0.004)} & {0.021 (0.002)} 
& {\bf 0.305 (0.022)} & {\bf 0.036 (0.009)} \\
\hline
PMF-5 (I) & {\bf 0.272 (0.004)} & {0.023 (0.001)}
& -- & --\\
\hline
KPMF-20 (D) & {0.275 (0.005)} & {0.031 (0.003)} 
& {0.306 (0.022)} & {0.035 (0.007)} \\
\hline
PMF-20 (I) & {0.273 (0.005)} & {0.023 (0.001)} 
& -- & --\\
\noalign{\smallskip}\hline
\end{tabular}
\end{table}

\begin{figure}[!th]
    \begin{center}
    \subfigure[Flixster (known users) recall$@k$]{
    \includegraphics[width=0.45\textwidth]{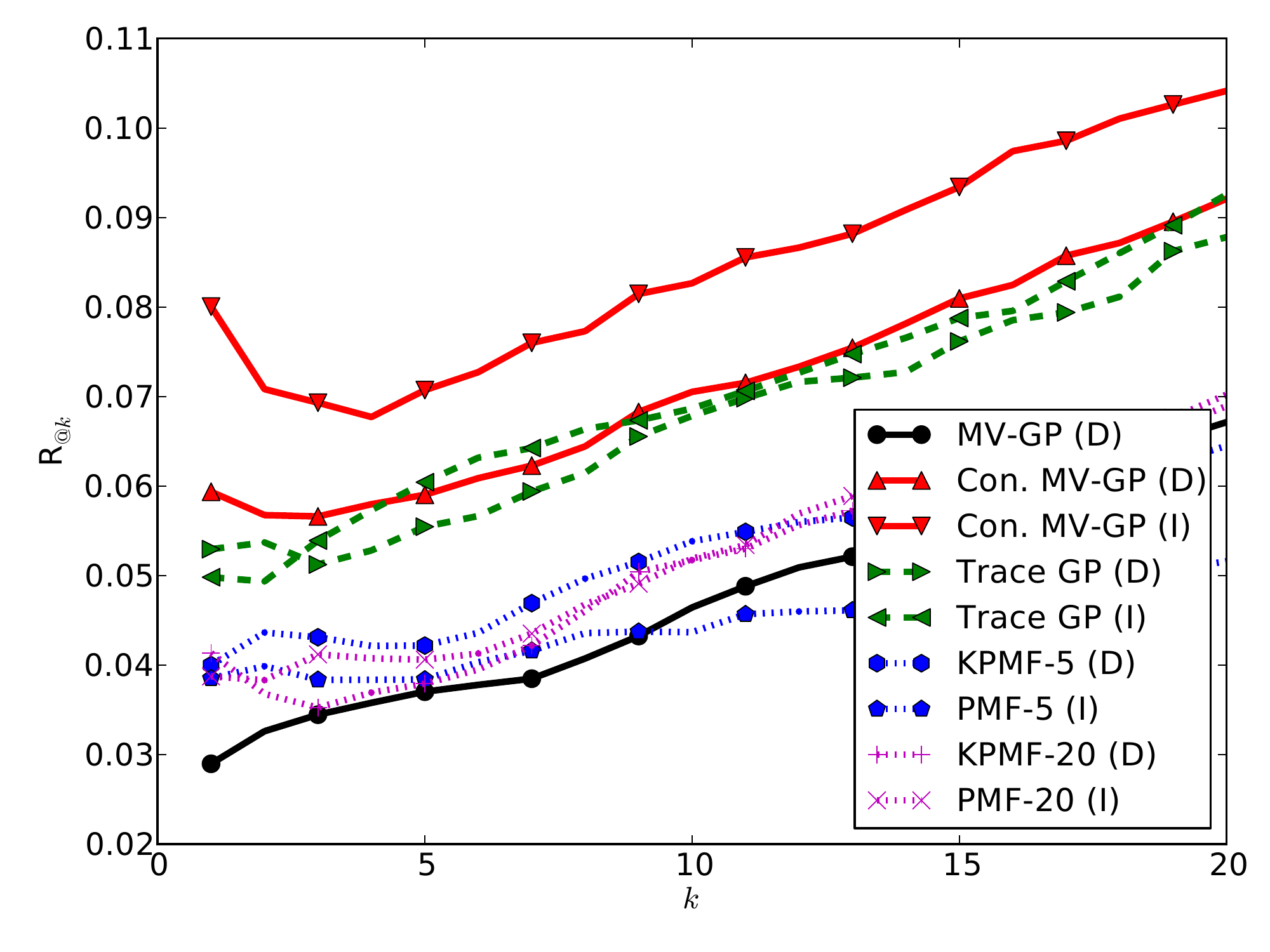}
    \label{fig:flixtr_mat_all}
	}%
	\quad
    \subfigure[Flixster (new users) recall$@k$]{
    \includegraphics[width=0.45\textwidth]{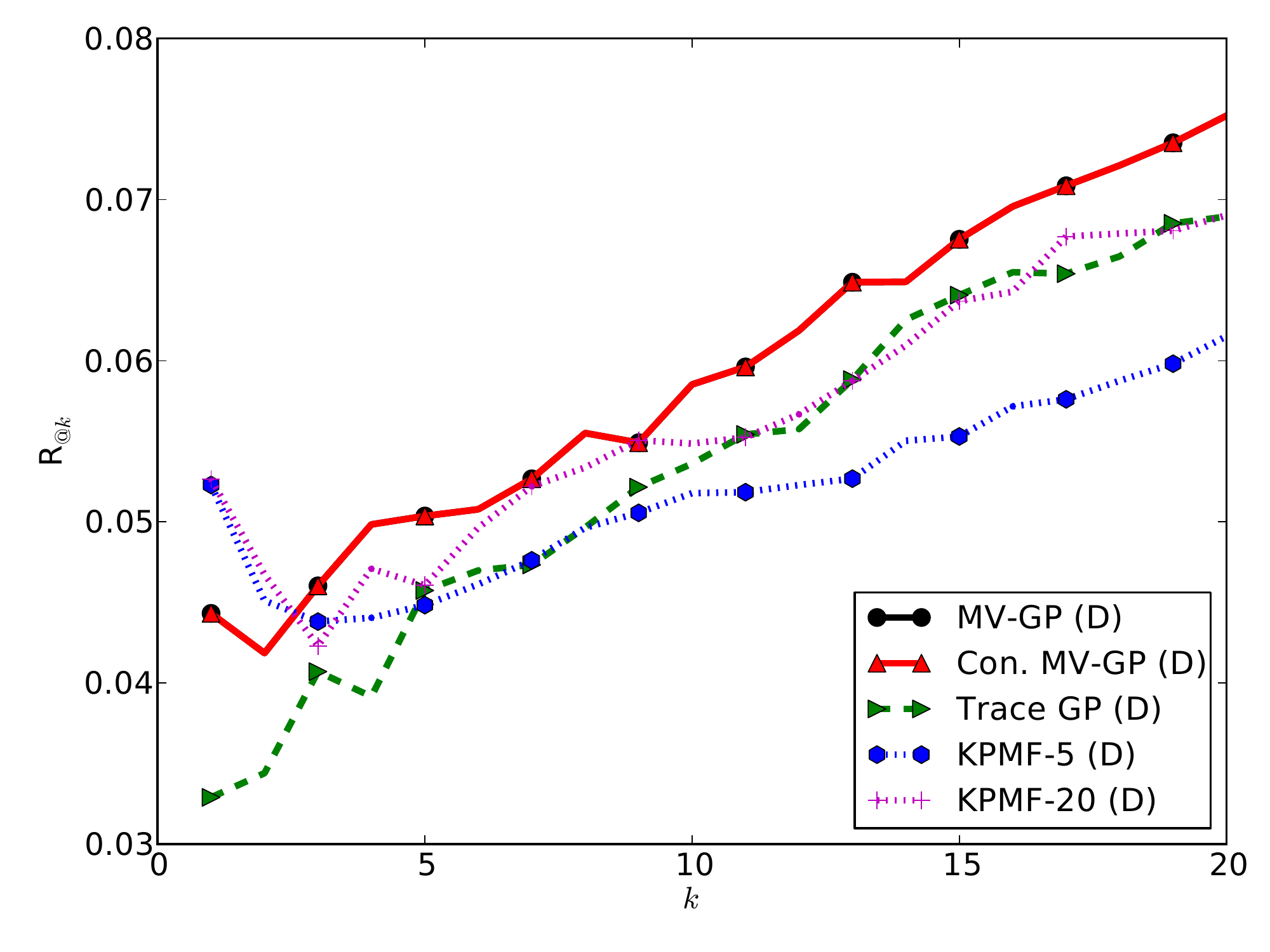}
    \label{fig:flixtr_cold_all}
	}

    \subfigure[Epinions (known users) recall$@k$]{
    \includegraphics[width=0.45\textwidth]{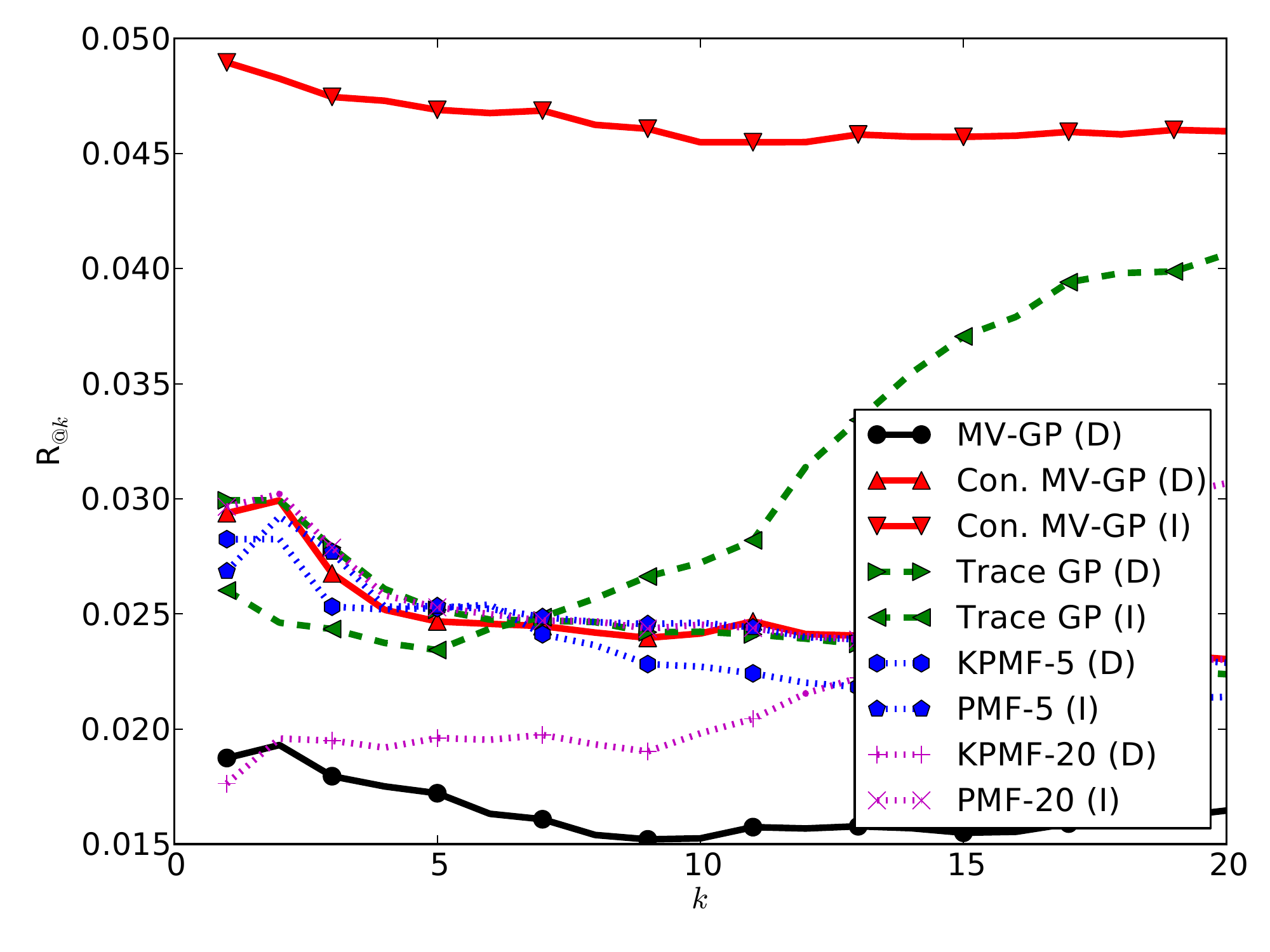}
    \label{fig:epinions_mat_all}
} \quad
    \subfigure[Epinions (new users) recall$@k$]{
    \includegraphics[width=0.45\textwidth]{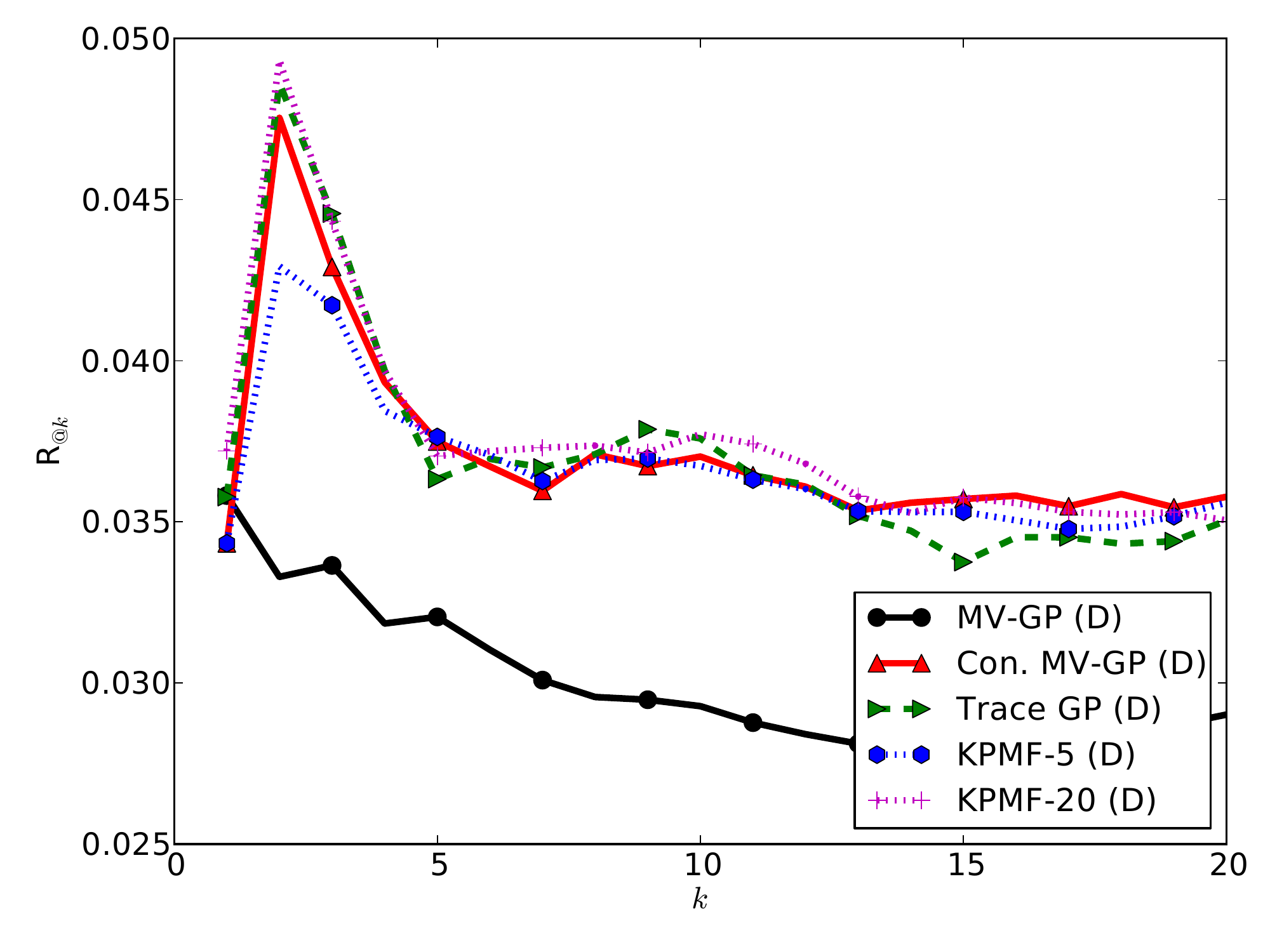}
    \label{fig:epinions_cold_all}
}
	\end{center}
    \caption{Performance results on recommender systems datasets. Recall $@k=1,
    2, \ldots, 20$ for known users (left) and new users (right).
    The prior covariances and constraints have the largest effect for very
    sparse data. (I): Identity prior covariance, (D): Diffusion prior
    covariance. Con. MV-GP outperforms KPMF \citep{zhou12}, which has been shown
    to outperform PMF \citep{ruslan08} and SoRec
    \citep{ma08}.}\label{fig:movie_expts}
\end{figure}

We present five fold cross validation performance for in matrix and new user
predictions on both Flixster and Epinions datasets.  We found that the model
that selected using $\rmse$ as the validation metric did not always perform best
in terms of recall (and vice versa). This matches the results by other
researchers \citep{steck10a, steck10b}. Hence, we performed cross validation
separately for $\rmse$ and $\recall{20}$. The results on the Flixster dataset
are shown in \tabref{tab:flixtr_mat_comb}. For known users, we found that the
tested models performed similarly in terms of RMSE, but the proposed models (Con
MV-GP, Trace GP) significantly outperformed the matrix factorization baselines
(KPMF, PMF) in terms of recall. These results are further highlighted in the
$\recall{k}$ performance as shown in \figref{fig:flixtr_mat_all}. The results
were often equivalent for new user predictions \figref{fig:flixtr_cold_all}.
Thus our experimental results suggest that the proposed models are more accurate
in terms of ranking while retaining competitive regression performance.

The RMSE and $\recall{20}$ performance on the Epinions dataset is shown in
\tabref{tab:epinions_mat_comb}. Our results here mirror the results on the
Flixster dataset. Our experiments show similar RMSE performance for all models,
and a significant gain in performance in terms of $\recall{20}$ for the proposed
constrained approach for known users. A similar trend is also highlighted in
\figref{fig:epinions_mat_all}. Con. MV-GP, Trace GP and KPMF perform similarly
when tested on new users as shown in \figref{fig:epinions_cold_all} with a
slight performance improvement for Con. MV-GP. Comparing the Bayesian MV-GP to
its constrained variant clearly shows the utility of the nuclear norm constraint
in both recommender systems datasets. In all, our results suggest that the
nuclear norm constrained MV-GP is effective for regression \emph{and} for
ranking in recommender systems.

\section{Conclusion}\label{sec:conclude}
This paper introduces a novel approach for the predictive modeling of low rank
transposable data with the matrix-variate Gaussian process.
The low rank is achieved using a nuclear norm constrained inference; recovering
a mean function of low rank. We showed that inference for the Gaussian process
with the nuclear norm constraint is convex. The proposed approach was applied to
the disease-gene association task and to the recommender system task. The
proposed model was effective for regression and for ranking with highly
imbalanced data, and performed at least as well as (and often significantly
better than) state of the art domain specific baseline models.

Recent work \citep{yu13} characterizing necessary and sufficient conditions for
the existence of a representer theorem points to the potential scope of the
constrained inference approach combined with nonparametric processes. Thus, we
plan to explore other constraint sets in addition to the nuclear norm constraint
explored here. We are also interested in exploring covariance constraints as
outlined for Gaussian distributions in \citet{koyejo13} applied to nonparametric
processes. We are interested in applications of nonparametric constrained
Bayesian inference to more complicated models beyond Gaussian distributions.
Finally, we intend to explore the biological implications of these constrained
disease gene association results in collaboration with domain experts.

\section*{Acknowledgements}
Authors acknowledge support from NSF grant IIS 1016614. We also thank
U.~Martin Blom and Edward Marcotte for providing the OMIM data set. The authors
thank the anonymous reviewers for insightful comments that helped to improve this
manuscript.

\appendix
\section{Constrained Bayesian Inference}\label{sec:con_proof}
\citet{altun06} studied the constrained inference approach when the constraint
set is a norm ball $\cC = \{ \bc\, \vert \, \| \bc - \bb\| \le \eps\}$.
They showed that one can apply the Fenchel duality theory to solve
\eqref{eq:conbayes} subject to such norm constraints and \citet{zhu12} extended
their approach to more general convex constraint sets. More recently
\citet{koyejo13} showed that the constrained Bayesian inference problem
satisfied a representer theorem in terms of exponential family distributions
under weak conditions. The discussion in this section follows the approach of
\citet{koyejo13}.

Let $\cX$ be a Banach space and $\cX^*$ be its dual space. The Legendre-Fenchel
transformation (convex conjugate) of a function $f: \cX \mapsto [-\infty,
+\infty]$ is $f^*:
\cX^* \mapsto [-\infty, +\infty]$ where $f^*(x^*) = \underset{x \in \cX}{\sup}\{
\innere{x}{x^*} - f(x)\}$. Further details on Fenchel duality may be found in
\citep{borwein05}.

Let $g(\cdot)$ denote a regularization function defined to match the properties
of the constraint set $\sfC$. For instance, we may define $g(\cdot)$ as an
indicator function of set membership in $\sfC$ or a soft penalty on the set
membership. The following theorem characterizes the solution of
\eqref{eq:conbayes_post} when $\sfC$ is convex.
\begin{theorem}[\citet{zhu12}] \label{theorem:dual}
Let $g$ be a convex function and denote its Legendre-Fenchel conjugate by $g^*$,
\begin{align} 
&\underset{q \in \cP}{\min}\; 
\kl{q(z)}{p(z)} + g(\ev{q}{\bgamma(z)}) \label{eq:primal}\\
=\, &\underset{\bkappa}{\max}\;  - \log \int_z p(z) 
\exp(\innere{\bkappa}{\bgamma(z)}) dz - g^*(-\bkappa) \label{eq:dual}
\end{align}
and the unique solution is given by $q_{*}(z) =
p(z)\exp(\innere{ (\kappa_{*}) }{\bgamma(z)} - \Lambda_{\bkappa_{*}} )$ where
$\bkappa_{*}$ is the solution of the finite dimensional dual optimization
\eqref{eq:dual} and $\Lambda_{\bkappa_{*}}$ ensures normalization.
\end{theorem}

Solving the resulting dual optimization \eqref{eq:dual} is often challenging. An
alternative primal approach is to separate the problem into two parts. First,
define the parametric form of the optimizing postdata density, then directly
optimize over that parametric family. Unlike the dual approach, the proposed
primal approach does not require convexity  of the constraint set.
However, both approaches require that a solution exists i.e. the set of
densities that satisfy \eqref{eq:conbayes_post} is not empty. For completeness,
we present the details of the solution.

Denote the constraint set subject to equality constraints as
$\cE_{\bc} = \{ q \in \cP \, | \, \ev{q}{\bgamma(z)} = \bc\}$. 
The constrained Bayes optimization problem can be written as:
\begin{equation}
\underset{\bc \in \sfC}{\min} \left[
\underset{q \in \cE_{\bc}}{\min}\;
\kl{q(z)}{p(z|y)}
\right], \label{eq:prime}
\end{equation}
which requires the solution of an inner optimization:
\begin{equation}
q_{\bc}(z) = \underset{q \in \cE_{\bc}}{\arg \min}\;
\kl{q(z)}{p(z|y)}. \label{eq:qx}
\end{equation}

Let $\sfA \subset \sfC$ be the set of points where the minimizer of
\eqref{eq:qx} is achievable. We can associate a density function $q_{\bc}(z)$
with every element $\bc \in \sfA$. The \emph{feasible set} is characterized by
the set of densities $\cS = \{q_{\bc}(z)\, \vert \, \bc \in \sfA \}$. 
The following proposition is a direct consequence of \thref{theorem:dual} and is
stated without proof.
\begin{proposition}[\citet{koyejo13}]\label{prop:equal}
For any $\bc \in \sfA$, the unique minimizer of \eqref{eq:qx} is given by:
$q_{\bc}(z) = p(z|y)\exp(\innere{ \bkappa_{\bc} }{\bgamma(z)} -
\Lambda_{\bkappa_{\bc}} )$ where $\bkappa_{\bc}$ is the solution of the finite
dimensional dual optimization \eqref{eq:dual} with the constraint set $\cC' =
\{\ev{q}{\bgamma(z)} = \bc\}$ and $\Lambda_{\bkappa_{\bc}}$ ensures
normalization.
\end{proposition}

\begin{theorem}[\citet{koyejo13}] \label{theorem:primal}
Let $\cS = \{q_{\bc} \, | \, \bc \in \sfA \}$ denote the feasible set of
\eqref{eq:qx}. The postdata density given by the minimizer of
\eqref{eq:conbayes_post} is the solution of:
\begin{equation*}
q_{*}(z) = \underset{q \in \cS }{\arg\min}\;  \kl{q(z)}{p(z|y)}
\end{equation*}
and the solution is given by $q_*(z) = q_{\ba}(z) $ for the optimal $\ba \in
\sfA$ with $ q_*(z) = p(z|y)\exp(\innere{(\bkappa_{\ba})}{\bkappa(z)} -
\Lambda_{\bkappa_{\ba}} )$ where $\bkappa_{\ba}$ is the solution of the finite
dimensional dual optimization \eqref{eq:dual} with the constraint set $\cC' =
\{\ev{q}{\bgamma(z)} = \ba\}$ and $\Lambda_{\bkappa_{\ba}}$ ensures
normalization.
\end{theorem}

The key insight from \propref{prop:equal} is that the solution of \eqref{eq:qx}
fully specifies the parametric form of the density. In other words, all the
members of the set $\cS = \{q_{\bc} \, | \, \bc \in \sfA \}$ have the same
parametric form with $q_{\bc} = f_{\btheta_{\bc}}(z)$ is determined by the
choice of $\bc$.
Note that all $\btheta \in \bTheta$ where $\bTheta$ is the constraint set of the
parametric distribution family specified by $f$. The existence of this
parameterized family follows from \thref{theorem:primal}.
\begin{corollary} \label{corr:parametric}
The postdata density given by the minimizer of
\eqref{eq:conbayes_post} is given by $q_*(z) = f_{\btheta_*}(z)$ where
$\btheta_*$ is the solution of:
\begin{equation*}
\btheta_* = \underset{\btheta \in \bTheta}{\arg\min}\; \bigg[  
\kl{f_{\btheta}(z)}{p(z|y)} \,
\sut \; \ev{f_{\btheta}}{\bgamma(z)} \in \cC
\bigg].
\end{equation*}
\end{corollary}
The expectation $\ev{f_{\btheta}}{\bgamma(z)}$ will be a fixed function
of $\btheta$ depending on the specific parametric family. Hence \corref{corr:parametric}
becomes a finite dimensional constrained optimization over $\btheta$.
\corref{corr:parametric} suggests the following recipe for constrained Bayesian
inference. First, \propref{prop:equal} is applied to specify the parametric form
of $q_{*}$, then \corref{corr:parametric} is applied to convert the variational
problem into a finite dimensional parametric optimization.

\section{Spectral Norms of Compact Operators}\label{sec:nuclearnorm}
Let $\hm$ denote the Hilbert space of functions induced by the row prior covariance
$\ckm$. Similarly, let $\hn$ denote the Hilbert space of  functions induced by
the column prior covariance $\ckn$. Let $\bx \in \hm$ and $\by \in \hn$ define (possibly
infinite dimensional) feature vectors.
The mean function the MV-GP is defined by a linear map $W: \hm \mapsto \hn$.
This is the bilinear form on $\hk=\hm \times \hn$ given by $\psi(m,n) =
\inner{\bx_m}{W \by_n}_{\hm}$.

Let $\cB$ denote the set of compact bilinear operators mapping $\hm \mapsto
\hn$. A compact operator $W \in \cB$ admits a spectral decomposition
\citep{abernethy09} with singular values given by $\{\xi_i(W)\}$. 
{\bf The nuclear norm} is given by the L1 norm on the spectrum of $W$:
\begin{equation} \label{eq:nuclear_norm}
\normht{\psi}{\hk} = \sum_{i=1}^D \xi_i(W)
\end{equation}
Another common regularizer is the induced {\bf Hilbert
norm} given by the L2 norm on the spectrum
of $W$:
\begin{equation} \label{eq:hilbert_norm}
\normhs{\psi}{\hk} = \sum_{i=1}^D \xi_i^2(W)
\end{equation}
Further details may be found in \citep{berlinet07}

Let $L(\psi, \by, \sfL)$ represent the loss function for a finite set of
training data points $\sfL \in \sfM \times \sfN$ and $Q(\psi)$ be a spectral
regularizer. We define the regularized risk functional:
\begin{equation*}
L(\psi, \by, \sfL) + \lambda Q(\psi)
\end{equation*}
where $\lambda \ge 0$ is the regularization constant. A representer theorem
exists, i.e., the function $\psi$ that optimizes the
regularized risk can be represented as a finite weighted
sum of the prior covariance functions evaluated on training data \citep{abernethy09}.
Hence, the optimizing function can be computed as:
\begin{align}\label{eq:representer}
\psi(m,n) 
&=\sum_{m' \in \sfM} \sum_{n' \in \sfN} \alpha_{m',n'} \,\ckm(m, m')\ckn(n,
n')\notag\\
&=\km(m)\bAlpha\kn(n)
\end{align}
where $\bAlpha \in \bbR^{M \times N}$ is a parameter matrix, $\km(m)$ is the prior covariance
matrix evaluated between $m$ and $m' \in \sfM$, i.e., the $m\th$ row of $\km$, and
$\kn(n)$ is the prior covariance matrix evaluated between $n$ and all $n' \in \sfN$.

\section{Parameter Estimation for the Nuclear Norm regularized
MV-GP}\label{sec:parameter} Like other Bayesian modeling approaches, the
constrained Bayesian inference procedure provides a mechanism for parameter
estimation. This is achieved by optimizing the cost function\footnote{Note that
the model evidence term must be added back in order to use the posterior form of
the constrained Bayesian inference cost function \eqref{eq:conbayes_post}.}
\eqref{eq:conbayes} with respect to the parameters.
The parameters of interest include the noise variance and the parameters of the
prior row and column covariance functions. The optimization for the noise
variance parameter is given by:
\begin{equation*}
\underset{\sigma^2}{\min}\;  
L \log \sigma^2 + \fracl{\sigma^2}
\left[
\sum_{m,n \in \sfL}(y\smn-\psi\smn)^2
+  \tr{\bS_{L}} \right]
\end{equation*}
This can be solved in closed form. The solution is given by:
\begin{equation}\label{eq:noise}
\sigma^2 = \fracl{L} \left[\sum_{m,n \in \sfL}(y\smn-\psi\smn)^2
+  \tr{\bS_{L}} \right]
\end{equation}
Similarly, we can solve for the parameters that define the prior covariance
functions.
Suppose the row covariance and column covariance have parametric forms $\km(\rho)$ and
$\kn(\tau)$ respectively. Let $\bC(\rho, \tau) = \kn(\tau) \kron \km(\rho)$
represent the joint prior covariance. We can select the covariance parameters by
optimizing \eqref{eq:conbayes} as:
\begin{equation*}
J(\rho, \tau)=
\underset{\rho, \tau}{\min} \; \;
\half \log | \bC(\rho, \tau) |
+ \half \trans{\bpsi}\inv{\bC(\rho, \tau)}\bpsi 
+ \half \text{tr}\bigg(\inv{\bC(\rho,
\tau)}\bS\bigg)
\end{equation*}
The gradient with respect to $\rho$ is given by:
\begin{gather*}
\pd{J(\rho, \tau)}{\rho}=
\frac{N}{2} \tr{ \inv{\km(\rho)}\pd{\km(\rho)}{\rho}  }
- \half \trans{\bpsi}\inv{\bC(\rho, \tau)} \D{\bC(\rho, \tau)}{\rho}\inv{\bC(\rho, \tau)}\bpsi 
\\
- \half \tr{\inv{\bC(\rho, \tau)} \D{\bC(\rho, \tau)}{\rho}\inv{\bC(\rho,
\tau)}\bS}
\end{gather*}
where $\pd{\bC(\rho, \tau)}{\rho} = \kn \kron \pd{\km(\rho)}{\rho}$, and
$\pd{\km(\rho)}{\rho}$ is the element-wise gradient. This can be
simplified further by collecting terms, and similar gradients can be computed
with respect to $\tau$. See \citep[Chapter 5]{rasmussen05} for more details on
the closely related approach of Gaussian process covariance parameter selection by
marginal likelihood optimization. We note that the prior covariance
hyperparameters may be computationally challenging to optimize in
practice as the proposed updates require the storage and computation of large
covariance matrices. 

\bibliographystyle{plainnat}      

\end{document}